\documentclass{article}

    \PassOptionsToPackage{numbers, sort&compress}{natbib}

\usepackage[preprint]{neurips_2025}

\usepackage[utf8]{inputenc} 
\usepackage[T1]{fontenc}    
\usepackage{hyperref}       
\usepackage{url}            
\usepackage{booktabs}       
\usepackage{amsfonts}       
\usepackage{nicefrac}       
\usepackage{microtype}      
\usepackage{xcolor}         
\usepackage{amsmath}
\usepackage{graphicx}

\title{SPIEDiff: robust learning of long-time macroscopic dynamics from short-time particle simulations with quantified epistemic uncertainty}

\author{%
  Zequn He \\
  Department of Mechanical Engineering and Applied Mechanics\\
  University of Pennsylvania\\
  Philadelphia, PA 19104 \\  \texttt{hezequn@seas.upenn.edu} \\
  \AND
  Celia Reina\thanks{Corresponding author} \\
  Department of Mechanical Engineering and Applied Mechanics\\
  University of Pennsylvania\\
  Philadelphia, PA 19104 \\
  \texttt{creina@seas.upenn.edu} \\
}

\begin{document}

\maketitle

\begin{abstract}
The data-driven discovery of long-time macroscopic dynamics and thermodynamics of dissipative systems with particle fidelity is hampered by significant obstacles. These include the strong time-scale limitations inherent to particle simulations, the non-uniqueness of the thermodynamic potentials and operators from given macroscopic dynamics, and the need for efficient uncertainty quantification. This paper introduces Statistical-Physics Informed Epistemic Diffusion Models (SPIEDiff), 
a machine learning framework designed to overcome these limitations in the context of purely dissipative systems by leveraging statistical physics, conditional diffusion models, and epinets. We evaluate the proposed framework on stochastic Arrhenius particle processes and demonstrate that SPIEDiff can accurately uncover both thermodynamics and kinetics, while enabling reliable long-time macroscopic predictions using only short-time particle simulation data. 
SPIEDiff can deliver accurate predictions with quantified uncertainty in minutes, drastically reducing the computational demand compared to direct particle simulations, which would take days or years in the examples considered. 
Overall, SPIEDiff offers a robust and trustworthy pathway for the data-driven discovery of thermodynamic models.
\end{abstract}

\section{Introduction}
\label{sec:intro}
Predicting the long-time evolution of dissipative, irreversible processes is vital across science and engineering, impacting applications from materials design to chemical reaction modeling \cite{simo2006computational,moeendarbary2009dissipative, dattagupta2013dissipative, danielewicz1985dissipative}. Although the fundamental description often resides at the level of interacting particles, direct particle simulations are typically computationally intractable to capture macroscopic phenomena. This limitation creates a critical need for coarse-graining methodologies. Extensive efforts have been made in pursuit of this goal through multiscale computational techniques \cite{tadmor2011modeling}, analytical approaches \cite{zwanzig2001nonequilibrium, givon2004extracting}, and increasingly, machine learning and data-driven strategies \cite{raissi2019physics, brunton2022data}. Yet, these approaches are often trapped in a compromise between physical fidelity and computational efficiency, and rarely quantify the uncertainty of the coarse-grained models and predictions.

To ensure that the derived macroscopic models are not merely phenomenological but also physically sound, they must embody fundamental thermodynamic principles. The General Equation for Non-Equilibrium Reversible-Irreversible Coupling (GENERIC) formalism \cite{grmela1997dynamics, ottinger1997dynamics, ottingerbeyond} offers a robust structure for developing such thermodynamically consistent evolution equations. For a closed system described by state variables $z(x,t)$, GENERIC postulates that the evolution of $z$ is governed by
\begin{equation}
\label{eq:generic}
\frac{\partial z}{\partial t}=\mathcal{L}_z \frac{\delta E[z]}{\delta z}+\mathcal{M}_z \frac{\delta S[z]}{\delta z},
\end{equation}  
where $E[z]$ and $S[z]$ are the total energy and entropy functionals, respectively, $\mathcal{L}_{z}$ is an antisymmetric Poisson operator satisfying the Jacobi identity, and
$\mathcal{M}_{z}$ is a symmetric, positive semi‐definite operator, responsible for the irreversible, entropy-producing part of the dynamics. Furthermore, the degeneracy conditions $\mathcal{L}_z \frac{\delta S}{\delta z}=\mathcal{M}_z \frac{\delta E}{\delta z}=0$ must hold to ensure 
that the system's evolution obeys conservation of energy and exhibits non-decreasing entropy, in full accordance with the 
laws of thermodynamics.

\begin{figure}[htbp]
    \centering
    \includegraphics[width=1.0\linewidth]{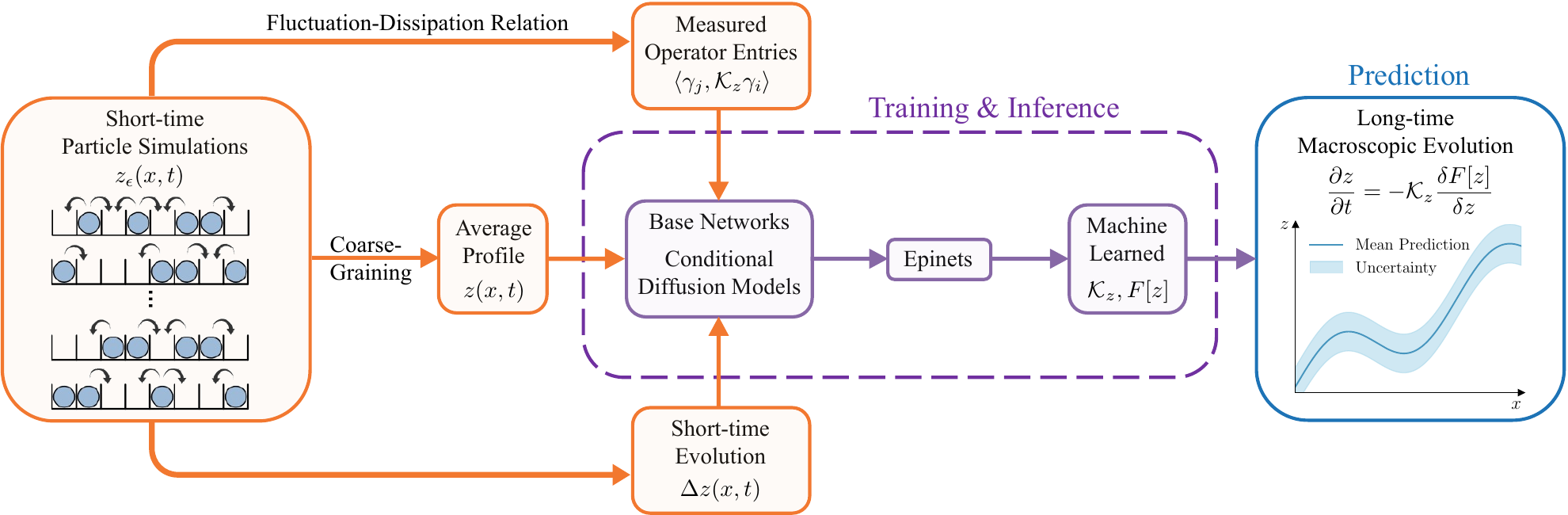}
    \caption{Schematic overview of the proposed framework for purely dissipative isothermal processes $\partial z/\partial t =-\mathcal{K}_z \delta F/\delta z$ (or analogously, $\partial z/\partial t=\mathcal{M}_z \delta S/\delta z$ in closed systems). Data collected from particle simulations $z_{\epsilon}(x,t)$ serve as inputs, specifically: the coarse-grained field $z(x,t)=\mathbb{E}\left[z_\epsilon\right]$, its short-time evolution $\Delta z(x,t)$, and the discretized dissipative operator entries $\left\langle\gamma_j, \mathcal{K}_z \gamma_i\right\rangle$, using finite element basis $\{\gamma\}$, measured via fluctuation-dissipation relations. These inputs are used in the SPIEDiff (conditional diffusion models augmented with epinets). The framework is trained to learn the underlying thermodynamic structure, i.e., the dissipative operator $\mathcal{K}_z$ and the free energy functional $F[z]$, which are later used to discover the macroscopic evolution equations with quantified epistemic uncertainty.}
    \label{fig:spiediffOverview}
\end{figure}

Various scientific machine learning (SciML) frameworks have been developed to learn the key components of GENERIC equations \cite{hernandez2021structure, lee2021machine, zhang2022gfinns, gruber2024efficiently, huang2022variational}. However, this problem is often non-unique from purely macroscopic observables, meaning that multiple operators $\mathcal{L}_z$, $\mathcal{M}_z$, and functionals $E[z]$ and $S[z]$ can lead to the same coarse-grained dynamics. This severely limits the physics learned from the problem at hand, as such quantities are unique to each particle system. Recently, Statistical-Physics Informed Neural Networks (Stat-PINNs) \cite{huang2025statistical} address this problem for purely dissipative dynamics (i.e., $\frac{\partial z}{\partial t}=\mathcal{M}_z \frac{\delta S[z]}{\delta z}$), by further incorporating statistical mechanics relations within the learning strategy. In particular, fluctuation-dissipation relations \cite{li2019harnessing} 
are used to directly compute the dissipative operator $\mathcal{M}_z$ from the thermal fluctuations of $z(t)$, which are quantifiable observables in particle simulations. Furthermore, Stat-PINNs learn the unique thermodynamic structure of the dynamics from short-time particle simulations (trivially parallelizable), as opposed to long-time particle simulations, which are typically computationally intractable. Despite the success of Stat-PINNs, their practical application can encounter challenges regarding the robustness of the learned components when data obtained from particle simulations is limited and noisy (a common scenario). Moreover, performing efficient epistemic (model) uncertainty quantification (UQ) for both the discovered dissipative operator and the thermodynamics potential, as well as for the resulting continuum predictions, remains a key requirement for building a trustworthy data-driven model. Motivated by these, we propose SPIEDiff (illustrated in Fig.~\ref{fig:spiediffOverview}) in this work.

\textbf{Primary contributions:}

\textbf{Enhanced robustness in learning the thermodynamic structure and ensuing dynamics.} Deterministic models, such as multilayer perceptrons (MLPs) used in Stat-PINNs, learn point-to-point mappings, which can limit their robustness when trained on noisy/limited data. Generative models, in contrast, learn underlying data distributions, often providing greater robustness to such data imperfections. Therefore, we use conditional diffusion models \cite{ho2020denoising} as the backbones for SPIEDiff, which significantly increases the robustness of the learning strategy as compared to Stat-PINNs.

\textbf{Efficient epistemic uncertainty quantification.} A core contribution is SPIEDiff's efficient quantification of epistemic uncertainty for learned thermodynamic components and subsequent macroscopic predictions. This is achieved by equipping the conditional diffusion models with epinets \cite{osband2023epistemic}. This design only requires minimal computational cost compared to traditional ensemble methods \cite{lakshminarayanan2017simple} or Bayesian inference \cite{neal2012bayesian} for UQ.


\textbf{Demonstrated efficacy.} 
We validate SPIEDiff on various Arrhenius-type particle systems, including cases with long-range interactions, where the coarse-grained dynamics are analytically known, as well as with short-range interactions, where the accuracy of SPIEDiff is assessed against long-time particle simulations. 
In all cases, SPIEDiff accurately discovers the correct underlying thermodynamic structures and 
macroscopic dynamics at a low computational cost, 
while simultaneously providing reliable epistemic uncertainty bounds.

\section{Related works}
\label{sec:relatedWork}
\textbf{Machine learning strategies for GENERIC.} Machine learning approaches to uncover the GENERIC structure of macroscopic dynamics have rapidly advanced, starting with soft-constrained methods such as SPNNs \cite{hernandez2021structure}, where the degeneracy conditions are encoded in the loss function as a penalty term. While this permits flexibility in parameterizing the approximate operators, it does not strictly guarantee that the resulting models will satisfy the first and second laws of thermodynamics. Subsequent efforts have focused on hard-constrained methods that embed the GENERIC structure directly into the network architecture to provide stronger physical guarantees. For instance, GNODE \cite{lee2021machine} parametrizes the reversible and irreversible brackets using antisymmetric and symmetric tensor decompositions, respectively, and guarantees the required degeneracy conditions by design, with no need for any extra penalty terms. GFINNs \cite{zhang2022gfinns} embed orthogonal projections onto the kernels of known or learnable operators directly within their network parameterizations to satisfy the required symmetries and degeneracy conditions. More recently, NMS \cite{gruber2024efficiently} implements parameterization of operators based on exterior calculus, providing provable guarantees and optimal quadratic scaling. Despite these advancements, methods relying solely on macroscopic average dynamics typically face the non-uniqueness issue in disentangling the operator and potential components, a challenge specifically addressed by Stat-PINNs \cite{huang2025statistical}, as has been discussed in Section \ref{sec:intro}. SPIEDiff incorporates key technical components originally developed in Stat-PINNs, which are later elaborated in Section \ref{sec:methods}.

\textbf{UQ in SciML.} Traditional UQ methods include Gaussian processes \cite{rasmussen2003gaussian}, Bayesian inference \cite{neal2012bayesian, kononenko1989bayesian, gelman1995bayesian}, polynomial chaos expansions \cite{najm2009uncertainty}, deep ensembles \cite{lakshminarayanan2017simple, fort2019deep}, and Monte Carlo dropout \cite{gal2016dropout}. Several of these techniques have been successfully integrated with PINNs, leading to approaches like B-PINNs \cite{yang2021b}, PC$^2$ \cite{novak2024physics}, and dropout-PINNs \cite{zhang2019quantifying}. Recent advancements in generative modeling, including VAEs \cite{kingma2013auto} and GANs \cite{goodfellow2014generative}, have also been applied to physics-informed learning \cite{zhong2023pi, shin2023physics, yang2019adversarial, daw2021pid, gao2022wasserstein}. While these generative approaches can naturally quantify uncertainty with a lower computational cost compared to the traditional UQ methods, the effective decomposition of total uncertainty into its aleatoric and epistemic sources remains challenging. Addressing these considerations, ENNs \cite{osband2023epistemic} represent a promising novel paradigm. ENNs typically augment a base predictive model with a flexible and lightweight network termed an epinet. This epinet is specifically trained to model the variation in the base network's outputs attributable to epistemic uncertainty, often with modest computational overhead. Such a method is starting to be explored for UQ in SciML applications, as seen in works like NEON \cite{guilhoto2024composite}, EVODMs \cite{he2025evodms}, and E-PINNs \cite{nair2025pinns}.

\section{Preliminaries}
\label{sec:prelim}
\textbf{Conditional DDPM:} Denoising Diffusion Probabilistic Models (DDPMs) \cite{ho2020denoising} learn data distributions and generate samples by reversing a forward process that gradually corrupts data samples into pure Gaussian noise. The forward process, shared by both the unconditional and conditional DDPMs, is defined as a Markov chain over $\Omega$ discrete timesteps. Starting with data samples $\mathbf{y}_0 \sim q\left(\mathbf{y}_0\right)$, noise is added sequentially via $q\left(\mathbf{y}_{1: \Omega} \mid \mathbf{y}_0\right)=\prod_{\omega=1}^\Omega q\left(\mathbf{y}_\omega \mid \mathbf{y}_{\omega-1}\right)$, with $q\left(\mathbf{y}_\omega \mid \mathbf{y}_{\omega-1}\right)=\mathcal{N}\left(\mathbf{y}_\omega; \sqrt{1-\beta_\omega} \mathbf{y}_{\omega-1}, \beta_\omega \mathbf{I}\right)$. Here, $\left\{\beta_\omega \in(0,1)\right\}_{\omega=1}^\Omega$ represents a predefined variance schedule controlling the noise level added at each step $\omega$. A useful property of this process is that the marginal distribution of $\mathbf{y}_\omega$ given $\mathbf{y}_0$ remains Gaussian and is analytically tractable: $q\left(\mathbf{y}_\omega \mid \mathbf{y}_0\right)=\mathcal{N}\left(\mathbf{y}_\omega; \sqrt{\bar{\alpha}_\omega} \mathbf{y}_0,\left(1-\bar{\alpha}_\omega\right) \mathbf{I}\right)$, where $\alpha_\omega=1-\beta_\omega$ and $\bar{\alpha}_\omega=\prod_{l=1}^\omega \alpha_l$. However, the reverse process is altered in the conditional setting. Specifically, the reverse transition $p_{\boldsymbol{\theta}}\left(\mathbf{y}_{\omega-1} \mid \mathbf{y}_\omega, \mathbf{x}\right)$ now depends on both the noisy state $\mathbf{y}_\omega$ and the conditioning variable $\mathbf{x}$. The reverse model becomes $p_{\boldsymbol{\theta}}\left(\mathbf{y}_{0: \Omega} \mid \mathbf{x}\right)=p\left(\mathbf{y}_\Omega\right) \prod_{\omega=1}^\Omega p_{\boldsymbol{\theta}}\left(\mathbf{y}_{\omega-1} \mid \mathbf{y}_\omega, \mathbf{x}\right)$, with $p_\Omega \sim \mathcal{N}(\mathbf{0}, \mathbf{I})$ and each transition parameterized as $p_{\boldsymbol{\theta}}\left(\mathbf{y}_{\omega-1} \mid \mathbf{y}_\omega, \mathbf{x}\right)=\mathcal{N}\left(\mathbf{y}_{\omega-1} \mid \boldsymbol{\mu}_{\boldsymbol{\theta}}\left(\mathbf{y}_\omega, \omega, \mathbf{x}\right), \sigma^2_q(\omega) \mathbf{I}\right)$. Here, $\boldsymbol{\mu}_{\boldsymbol{\theta}}\left(\mathbf{y}_\omega, \omega, \mathbf{x}\right)$ is predicted by a neural network, and $\sigma_q^2(\omega)$ is typically inherited from the forward process. The model is trained to minimize a variational upper bound on the negative conditional log-likelihood \cite{chan2024tutorial}.
In practice, when predicting the target ($\boldsymbol{\mu}_{\boldsymbol{\theta}}=\hat{\mathbf{y}}_{\boldsymbol{\theta}}$) a simplified training loss is used by sampling a timestep $\omega\sim \mathcal{U}(1, \Omega)$ and optimizing
\begin{equation}
\label{eq:cDDPMTrainLoss}
\boldsymbol{\theta}^*=\arg\min_{\boldsymbol{\theta}}\left\|\hat{\mathbf{y}}_{\boldsymbol{\theta}}\left(\mathbf{y}_\omega, \omega, \mathbf{x}\right)-\mathbf{y}_0\right\|^2 .
\end{equation}
During inference, when predicting targets, conditional DDPMs iteratively update according to
\begin{equation}
\label{eq:updateDDPM}
\mathbf{y}_{\omega-1}=\frac{\left(1-\bar{\alpha}_{\omega-1}\right) \sqrt{\alpha_\omega}}{1-\bar{\alpha}_\omega} \mathbf{y}_\omega+\frac{\left(1-\alpha_\omega\right) \sqrt{\bar{\alpha}_{\omega-1}}}{1-\bar{\alpha}_\omega} \hat{\mathbf{y}}_{\boldsymbol{\theta}}+\sigma_\omega \boldsymbol{\varepsilon}_\omega, \quad \boldsymbol{\varepsilon}_\omega \sim \mathcal{N}(0, \mathbf{I}) ,
\end{equation}
where the sampler variance is $\sigma_\omega^2=\frac{1-\alpha_{\omega-1}}{1-\alpha_\omega} \beta_\omega$.

\textbf{DDIM sampler:} A significant limitation of standard DDPMs is their slow sampling speed due to the sequential nature of the reverse process. Denoising diffusion implicit models (DDIMs) were proposed in \cite{song2020denoising} to overcome this by reformulating the generation as a non-Markovian process that utilizes the same trained neural networks as the corresponding DDPMs. The primary distinction lies in the modified update formula for the reverse process sampling
\begin{equation}
\mathbf{y}_{\omega-1}=\sqrt{\alpha_{\omega-1}} \hat{\mathbf{y}}_{\boldsymbol{\theta}}+\sqrt{\frac{1-\alpha_{\omega-1}-\sigma_\omega^2}{1-\alpha_\omega}}\left(\mathbf{y}_\omega-\sqrt{\alpha_\omega} \hat{\mathbf{y}}_{\boldsymbol{\theta}}\right)+\sigma_\omega \boldsymbol{\varepsilon}_\omega, \quad \boldsymbol{\varepsilon}_\omega \sim \mathcal{N}(0, \mathbf{I}) .
\end{equation}
Here, $\sigma_\omega^2=\eta^2 \frac{1-\alpha_{\omega-1}}{1-\alpha_\omega}\left(1-\frac{\alpha_\omega}{\alpha_{\omega-1}}\right)$ with $\eta \geq 0$ to be the hyperparameter that controls the stochasticity during sampling. The fully deterministic DDIM sampler ( $\eta=0$) generates samples much faster as fewer steps are required. Setting $\eta=1$ recovers the original DDPM sampler.

\textbf{Epistemic neural networks:} Epistemic Neural Networks (ENNs) \cite{osband2023epistemic} offer a modular approach to UQ. This is achieved by augmenting any base network (parameters $\boldsymbol{\zeta}$) with a lightweight epinet (parameters $\boldsymbol{\xi}$), which learns how the base network's outputs vary under epistemic uncertainty \cite{wen2021predictions, osband2022neural, osband2023epistemic}. The combined model (an ENN) generates joint predictions as
\begin{equation}
\label{eq:ennEqn}
y_{\boldsymbol{\theta}}(\mathbf{x}, \boldsymbol{\phi})
= \mu_{\boldsymbol{\zeta}}(\mathbf{x})
+ \sigma_{\boldsymbol{\xi}}\bigl(\mathrm{sg}\bigl[\mathbf{h}_{\boldsymbol{\zeta}}(\mathbf{x})\bigr], \boldsymbol{\phi}\bigr) ,     
\end{equation}
where $\boldsymbol{\theta}=(\boldsymbol{\zeta},\boldsymbol{\xi})$ are all trainable parameters and $\mu_{\boldsymbol{\zeta}}$ is the output of the base network. The features $\mathbf{h}_{\boldsymbol{\zeta}}(\mathbf{x})$ are obtained from intermediate representations of the base network (usually the final hidden layer concatenated with the inputs $\mathbf{x}$). The stop-gradient operator $\mathrm{sg}[\cdot]$ prevents gradients from following back into $\boldsymbol{\zeta}$ via this path. The epistemic indices $\boldsymbol{\phi}$ are drawn from a simple prior distribution $P_\Phi$, such as a uniform distribution or a standard Gaussian. The epinet from Eq.~\eqref{eq:ennEqn} contains a fixed prior network $\sigma^P$ and a learnable network $\sigma_{\boldsymbol{\xi}}^L$ such that $\sigma_{\boldsymbol{\xi}}=\sigma^P+\sigma_{\boldsymbol{\xi}}^L $. The prior network encodes prior beliefs about uncertainty and is implemented as an ensemble of randomly initialized networks: $\sigma^P\left(\mathbf{x}, \boldsymbol{\phi}\right)=\kappa \sum_{i=1}^{d_\Phi} p^i(x) \phi_i$, where $\kappa$ is a fixed prior scalar. The learnable network is defined as $\sigma_{\boldsymbol{\xi}}^L(\mathrm{sg}\left[\mathbf{h}_{\boldsymbol{\zeta}}(\mathbf{x})\right], \boldsymbol{\phi})=\mathrm{NN}_{\boldsymbol{\xi}}\left(\mathrm{sg}\left[\mathbf{h}_{\boldsymbol{\zeta}}(\mathbf{x})\right], \boldsymbol{\phi}\right)^T \boldsymbol{\phi}$. During the training, the parameters $\boldsymbol{\xi}$ are optimized to ensure the full ENN prediction accurately represents the desired posterior predictive distribution.  
   
\section{Methods}
\label{sec:methods}
Similarly to the setting of Stat-PINNs \cite{huang2025statistical}, we consider purely dissipative dynamics under isothermal conditions governed by $\frac{\partial z}{\partial t}=-\mathcal{K}_z \frac{\delta F[z]}{\delta z}$, where $\mathcal{K}_z$ is symmetric and positive semi-definite and $F[z]$ is the free energy functional with an associated density $f(z)$, i.e., $F[z]=\int f(z(x,t)) \, dx$. The methods discussed below are fully analogous for evolutions of the form $\frac{\partial z}{\partial t}=\mathcal{M}_z \frac{\delta S[z]}{\delta z}$ in closed systems.
\subsection{Discovery of the unique dissipative operator from fluctuations}
While macroscopic observations of the field $z$ alone are generally insufficient to uniquely determine the operators and thermodynamic potentials within the GENERIC structure, their fluctuations uniquely determine $\mathcal{K}_z$ through an infinite-dimensional fluctuation-dissipation relation \cite{li2019harnessing,huang2025statistical}. In particular, consider a basis of functions $\left\{\gamma_i(x)\right\}$ (e.g., a finite element basis) and denote by $z_\epsilon$ the measurement from particle simulations, necessarily stochastic, at a given level of characteristic particle volume $\epsilon$ (the continuum limit being $\epsilon \rightarrow 0$). Then, the discretized dissipative operator $\left\langle\gamma_j, \mathcal{K}_z \gamma_i\right\rangle$ ($\langle\cdot, \cdot\rangle$ denotes the $L^2$ inner product) can be computed as the covariation of the rescaled fluctuations, i.e.,
\begin{equation}
\label{eq:entryFluctuation}
\left\langle\gamma_j, \mathcal{K}_z \gamma_i\right\rangle=\lim _{h \searrow 0} \frac{1}{2 h} \mathbb{E}\left[\left(Y_{\gamma_i}\left(t_0+h\right)-Y_{\gamma_i}\left(t_0\right)\right) \cdot\left(Y_{\gamma_j}\left(t_0+h\right)-Y_{\gamma_j}\left(t_0\right)\right)\right].
\end{equation}
Here, the time step $h$ must be macroscopically small so that changes in $z=\mathbb{E}\left[z_\epsilon\right]$ are negligible, but microscopically large to capture particle fluctuations. $t_0$ is an initial time where the system has reached local equilibrium for the given macrostate $z$, and $Y_{\gamma}$ is the limit of $\left\langle z_\epsilon-z, \gamma\right\rangle / \sqrt{\epsilon}$. The detailed proof, which can be found in \cite{li2019harnessing}, relies on the fact that for finite but large number of particles (small $\epsilon$), $z_\epsilon$ is typically well-described by the stochastic PDE $\frac{\partial z_\epsilon}{\partial t}=-\mathcal{K}_{z_\epsilon} \frac{\delta F\left[z_\epsilon\right]}{\delta z_\epsilon}+\sqrt{2 \epsilon \mathcal{K}_{z_\epsilon}} \dot{W}_{x, t}$ interpreted in an Itô sense, where $\dot{W}_{x,t}$ is a space-time white noise.

\subsection{Discretization and structure-preserving parameterization}
We consider the structure-preserving parameterization of the discretized dissipative operator outlined in \cite{huang2025statistical}, which retains the symmetry and positive semi-definiteness of its infinite-dimensional counterpart in problems living in one spatial dimension, and discretized with piecewise-linear finite element shape functions. For a dissipative operator $\mathcal{K}_z$ that depends only on the local state $z(x,t)$ and acts locally on $Q=\delta F/\delta z$, its discretized version is a tridiagonal matrix, that is, $K_{ij}=\left\langle\gamma_j, \mathcal{K}_z \gamma_i\right\rangle \neq 0 \Longleftrightarrow j \in\{i-1, i, i+1\}$. Thus, only two functions are needed to fully characterize it: $K_0\left(\mathbf{Z}_i^0\right)=K_{ii}$ for the diagonal entries and $K_1\left(\mathbf{Z}_i^1\right)=K_{i+1, i}=K_{i, i+1}$ for the off-diagonal entries, where $\mathbf{Z}_i^0=\left(z_{i-1}, z_i, z_{i+1}\right)$ and $\mathbf{Z}_i^1=\left(z_i, z_{i+1}\right)$. These two functions are further related for conserved fields $z$ as
\begin{equation}
\label{eq:K0K1Relation}
K_1\left(\mathbf{Z}_{i-1}^1\right)+K_0\left(\mathbf{Z}_i^0\right)+K_1\left(\mathbf{Z}_i^1\right)=0 \quad \Longrightarrow \quad K_0\left(\mathbf{Z}_i^0\right)=-K_1\left(\mathbf{Z}_{i-1}^1\right)-K_1\left(\mathbf{Z}_i^1\right),
\end{equation}
which enables the calculation of $K_0$ from $K_1$. In this setting, the simple condition $K_1 \leq 0$ becomes sufficient to guarantee the positive semi-definiteness of the discretized operator for any field value $z(x,t)$. Detailed proofs can be found in \cite{huang2025statistical}.

\subsection{Statistical-physics informed epistemic diffusion models}
SPIEDiff are ENNs containing base networks and epinets to approximate $K_1$ and $f$ with quantified uncertainty, from which the full discretized operator, functional $F[z]=\int f dx$, and the ensuing dynamics may be obtained (see Fig.~\ref{fig:spiediffOverview}). The base networks are conditional DDPMs using MLPs as their backbones. The first conditional DDPM, denoted as $\mathrm{NN}_{K_1}$, is trained to approximate $K_1$ from fluctuation data (recall Eq.~\eqref{eq:entryFluctuation}) obtained from multiple realizations of short-time particle simulations.
The loss function used for training $\mathrm{NN}_{K_1}$ makes use of Eq.~\eqref{eq:K0K1Relation}, and is defined as
\begin{equation}
\begin{aligned}
\label{eq:lossFuncionK1Base}
\mathcal{L}^{\mathrm{Base}}_{K_1} = \mathbb{E}_{\substack{s \sim \mathcal{D}_{K_{1}}, \\ \omega \sim \mathcal{U}(1, \Omega)}}\Bigg[&\lambda^{K_1}_0 \left| -\hat{K}_1\left(\mathbf{Z}_{i^s-1}^{1(s)}, K_{1(\omega)}^{(s)}, \omega\right) - \hat{K}_1\left(\mathbf{Z}_{i^s}^{1(s)}, K_{1(\omega)}^{(s)}, \omega\right) - K_0^{(s)}\right|^2 \\
& + \lambda^{K_1}_1\left| \hat{K}_1\left(\mathbf{Z}_{i^s}^{1(s)}, K_{1(\omega)}^{(s)}, \omega\right) - K_1^{(s)}\right|^2 \Bigg] ,
\end{aligned}
\end{equation}
where $\hat{K}_1$ denotes the output of $\mathrm{NN}_{K_1}$ parameterized by $\boldsymbol{\zeta}_{K_1}$. $\lambda^{K_1}_0$ and $\lambda^{K_1}_1$ are the loss term weights which are set to $1$ in all cases for simplicity. $\mathcal{D}_{K_1}$ consists of $\left\{K_0^{(s)}, K_1^{(s)}, \mathbf{Z}_{i^s-1}^{1(s)}, \mathbf{Z}_{i^s}^{1(s)}\right\}_{s=1}^{N_K}$ with $N_K$ being the number of training data and $i^s$ being the spatial index for sample $s$. $\omega$ is the diffusion timestep and $K_{1(\omega)}^{(s)}$ is the noisy version of $K_1^{(s)}$ at that step. Note that we apply a transformation function $g(\cdot)$ to $\mathrm{NN}_{K_1}$ to ensure that the learned operator is positive semi-definite as discussed above (see Appendix \ref{sec:modelArch} for details). 


After learning the discretized dissipative operator, a second conditional DDPM ($\mathrm{NN}_{f}$), parameterized by $\boldsymbol{\zeta}_f$, is trained to learn the free energy density $f$. The training data is collected from the system's macroscopic evolution over a small timestep $\Delta t$. It is important to distinguish the timescales involved: this $\Delta t$ must be macroscopically small yet significantly large for changes in $z$ to be measurable, and it is hence much larger than the timescale $h$ previously used for estimating operator entries from fluctuations. The loss function is a physics-informed loss based on the residual of the discretized evolution equations $\sum_{i=1}^{N_\gamma}\left\langle\gamma_j, \gamma_i\right\rangle \dot{z}_i(t)= \sum_{i=1}^{N_\gamma}\left\langle\gamma_j, \mathcal{K}_z \gamma_i\right\rangle Q_i(t)$, $\forall j=1, \ldots, N_\gamma$. It reads
\begin{equation}
\begin{aligned}
\label{eq:lossFuncionFEBase}
\mathcal{L}^{\mathrm{Base}}_{f} = \mathbb{E}_{\substack{s \sim \mathcal{D}_{f}, \\ \omega \sim \mathcal{U}(1, \Omega)}}\Bigg[&\lambda^{f}_0 \left| \sum_i\left\langle\gamma_{j}, \gamma_i\right\rangle \frac{\Delta z_i^{(s)}}{\Delta t}+\sum_i\left\langle\gamma_{j}, \mathcal{K}_{z^{(s)}} \gamma_i\right\rangle \hat{\boldsymbol{Q}}\left(\mathbf{Z}_i^{f(s)}, \Upsilon^{(s)}_{\omega}, \omega\right) \right|^2 \\
& + \lambda^{f}_1\left| \hat{\Upsilon}\left(\mathbf{Z}_i^{f(s)}, \Upsilon^{(s)}_{\omega}, \omega\right) - \Upsilon^{(s)} \right|^2\Bigg] ,
\end{aligned}
\end{equation}
where $\mathcal{D}_f$ is the training dataset containing $\left\{\left\langle\gamma_{j}, \gamma_i\right\rangle \Delta z_i^{(s)} / \Delta t, \left\langle\gamma_{j}, \mathcal{K}_{z^{(s)}} \gamma_i\right\rangle, \mathbf{Z}_i^{f(s)}, \Upsilon^{(s)}\right\}_{s=1}^{N_f}$ for $N_f$ training data points. Here, $\left\langle\gamma_j, \mathcal{K}_{z^{(s)}} \gamma_i\right\rangle$ is determined using the pre-trained $\mathrm{NN}_{K_1}$. $\mathbf{Z}_i^{f(s)}$ is the vector containing the field value $z_i^{(s)}$ at node $i$ for sample $s$. $\Upsilon^{(s)}$ is the auxiliary quantity that serves as the data target in the forward and reverse processes of the diffusion model. 
In this study, $\Upsilon^{(s)}$ is chosen as $\sum_i\left\langle\gamma_{j}, \gamma_i\right\rangle \frac{\Delta z_i^{(s)}}{\Delta t}$, although other quantities like $\mathbf{Z}_i^{f(s)}$ could also be used. The choice of $\Upsilon^{(s)}$ will not impact the model performance significantly. We set $\lambda^{f}_0=\lambda^{f}_1=1$ in all experiments. 

Upon completion of the sequential training for the base networks, the base predictions $\mu_{\boldsymbol{\zeta}_{K_1}}$ and $\mu_{\boldsymbol{\zeta}_{f}}$, as well as the features $\mathbf{h}_{\boldsymbol{\zeta}_{K_1}}$ and $\mathbf{h}_{\boldsymbol{\zeta}_{f}}$ are generated using the DDIM sampler for fast sampling, and used for training the corresponding epinets. The primary objective of this work is the quantification of epistemic uncertainty. However, accessing training data completely devoid of aleatoric uncertainty (i.e., perfectly "clean" data) is impractical. To approximate epistemic uncertainty under such conditions, we employ knowledge distillation \cite{hinton2015distilling} for training $\mathrm{Epinet}_{K_1}$ (parameterized by $\boldsymbol{\xi}_{K_1}$) and $\mathrm{Epinet}_{f}$ (parameterized by $\boldsymbol{\xi}_{f}$). The teacher models are the pre-trained conditional DDPMs, $\mathrm{NN}_{K_1}$ and $\mathrm{NN}_{f}$, respectively, which provide target predictions: $\hat{K}_1^{(s)}$ and $\hat{f}^{(s)}$ for each sample $s$. Meanwhile, a target diagonal term $\hat{K}_0^{(s)}$ is derived from $\hat{K}_1^{(s)}$ via Eq.~\eqref{eq:K0K1Relation}. The SPIEDiff's prediction for the component $K_1$, given an input $\mathbf{Z}^1$ and sampled epistemic indices $\boldsymbol{\phi}_{K_1} \sim P_\Phi$ (e.g., a standard Gaussian), is $\tilde{K}_1\left(\mathbf{Z}^1, \boldsymbol{\phi}_{K_1}\right)=\mu_{\boldsymbol{\zeta}_{K_1}}\left(\mathbf{Z}^1\right)+\sigma_{\boldsymbol{\xi}_{K_1}}\left(\mathrm{sg}\left[\mathbf{h}_{\boldsymbol{\zeta}_{K_1}}\left(\mathbf{Z}^1\right)\right], \boldsymbol{\phi}_{K_1}\right)$. Here, $\mu_{\boldsymbol{\zeta}_{K_1}}\left(\mathbf{Z}^1\right)$ is the fixed base prediction from $\mathrm{NN}_{K_1}$ and $\sigma_{\boldsymbol{\xi}_{K_1}}\left(\mathrm{sg}\left[\mathbf{h}_{\boldsymbol{\zeta}_{K_1}}\left(\mathbf{Z}^1\right)\right], \boldsymbol{\phi}_{K_1}\right)$ is the output of $\mathrm{Epinet}_{K_1}$. $\tilde{f}(\mathbf{Z}^{f}, \phi_{f})$ is defined similarly. The loss function minimized to jointly train both epinets is 
\begin{equation}
\begin{aligned}
\label{eq:lossFunctionEpinets}
\mathcal{L}^{\mathrm{Epinets}} &= \frac{1}{M_{K_1}} \sum_{\boldsymbol{\phi}_{K_1} \in \Phi} \sum_{s \in \tilde{\mathcal{D}}_{K_1}} \Bigg[
\left| -\tilde{K}_1\left(\mathbf{Z}_{i^s-1}^{1(s)}, \boldsymbol{\phi}_{K_1}\right) - \tilde{K}_1\left(\mathbf{Z}_{i^s}^{1(s)}, \boldsymbol{\phi}_{K_1}\right)-\hat{K}_0^{(s)}\right|^2 \\
& + \left|\tilde{K}_1\left(\mathbf{Z}_{i^s}^{1(s)}, \boldsymbol{\phi}_{K_1}\right)-\hat{K}_1^{(s)}\right|^2 \Bigg] + \frac{1}{M_f} \sum_{\boldsymbol{\phi}_{f} \in \Phi} \sum_{s \in \tilde{\mathcal{D}}_{f}} \Bigg[\left|\tilde{f}\left(\mathbf{Z}_{i^s}^{f(s)}, \boldsymbol{\phi}_{f}\right)-\hat{f}^{(s)}\right|^2 \Bigg],
\end{aligned}
\end{equation}
where $\tilde{D}_{K_1}$ and $\tilde{D}_f$ are the training datasets. $M_{K_1}=\left|\Phi\right| \cdot |\tilde{\mathcal{D}}_{K_1}|$ and $M_f=\left|\Phi\right| \cdot |\tilde{\mathcal{D}}_{f}|$ are the prefactors. Note that since the loss weights are all set to $1$ here, they are omitted. The size of the epistemic indices used for each epinet could be different, but we keep them the same but generated from different random seeds in our experiments for simplicity. Finally, the learned dissipative operator $\mathcal{K}_z$ and free energy functional $F[z]$ fully determine the discretized evolution equations. These are then solved using a fourth-order Runge-Kutta (RK4) method to generate continuum predictions. Comprehensive details of the SPIEDiff training procedure, specific network architectures, and all hyperparameter settings are provided in Appendix \ref{sec:modelArchTrain}.

\section{Experiments}
\label{sec:experiments}
\textbf{Arrhenius-type interacting particle systems.}
We consider a system of interacting particles confined to a one-dimensional lattice with lattice spacing $\epsilon$. Each lattice site $x$ can be either occupied, denoted by an occupation number $\eta(x)=1$, or empty, $\eta(x)=0$. The particles evolve via a stochastic jumping process, restricted to hops between nearest-neighbor sites. The rate for a particle to jump from an occupied site $x$ to an empty adjacent site $y$ is given by $p(x \rightarrow y)=d\, \eta(x)[1-\eta(y)] \exp \left[-\beta\left(\hat{U}_0+\sum_{\chi \neq x} \hat{J}(x-\chi) \eta(\chi)\right)\right]$. In this Arrhenius-type expression, $d$ is the jumping frequency, $\hat{U}_0$ the binding energy, and $\hat{J}(x-\chi)$ the interaction potential between a particle at site $x$ and another particle located at site $\chi$. Training data used in this section is generated using kinetic Monte Carlo (KMC) simulations of a 1D particle system with 2000 lattice sites (and 25 shape functions $\{\gamma\}$) initiated from 28 distinct initial density profiles $\rho(x,t)$ (listed in Appendix \ref{sec:dataGenPrepare}). Operator entries are estimated from fluctuations using $10$ time intervals ($h$) across $R=10^4$ realizations per initial profile (effective $R_K=10^5$). Free energy density is learned using the average macroscopic change over $\Delta t \gg h$ from $R=10^4$ realizations per profile. 
Details of the KMC simulation data are provided in Appendix \ref{sec:dataGenPrepare}. 
The learned off-diagonal term $K_1(\rho,_i\rho_{i+1})$ is equivalently expressed as $\bar{K}_1\left(\rho_{i+\frac{1}{2}},\left.\nabla \rho\right|_{i+\frac{1}{2}}\right)$, and the learned free energy density is calibrated and denoted as $\bar{f}$ (see Appendix \ref{sec:inferenceContPrediction}).


\textbf{Evaluation.}
Although the underlying Arrhenius diffusion model is conceptually straightforward, its dynamics and thermodynamic structure in the continuum limit are only known for the case of long-range interactions \cite{vlachos2000derivation} (this analytical model, here denoted as LRM, is given in Appendix \ref{sec:longRangeModel}). For this case, we will quantify the relative $L^2$ error between the proposed SPIEDiff model and the LRM.
For the cases of short-range interactions, where the LRM fails, the effectiveness of SPIEDiff will be evaluated against KMC simulations.

\begin{figure}[htbp]
    \centering
    \includegraphics[width=1.0\linewidth]{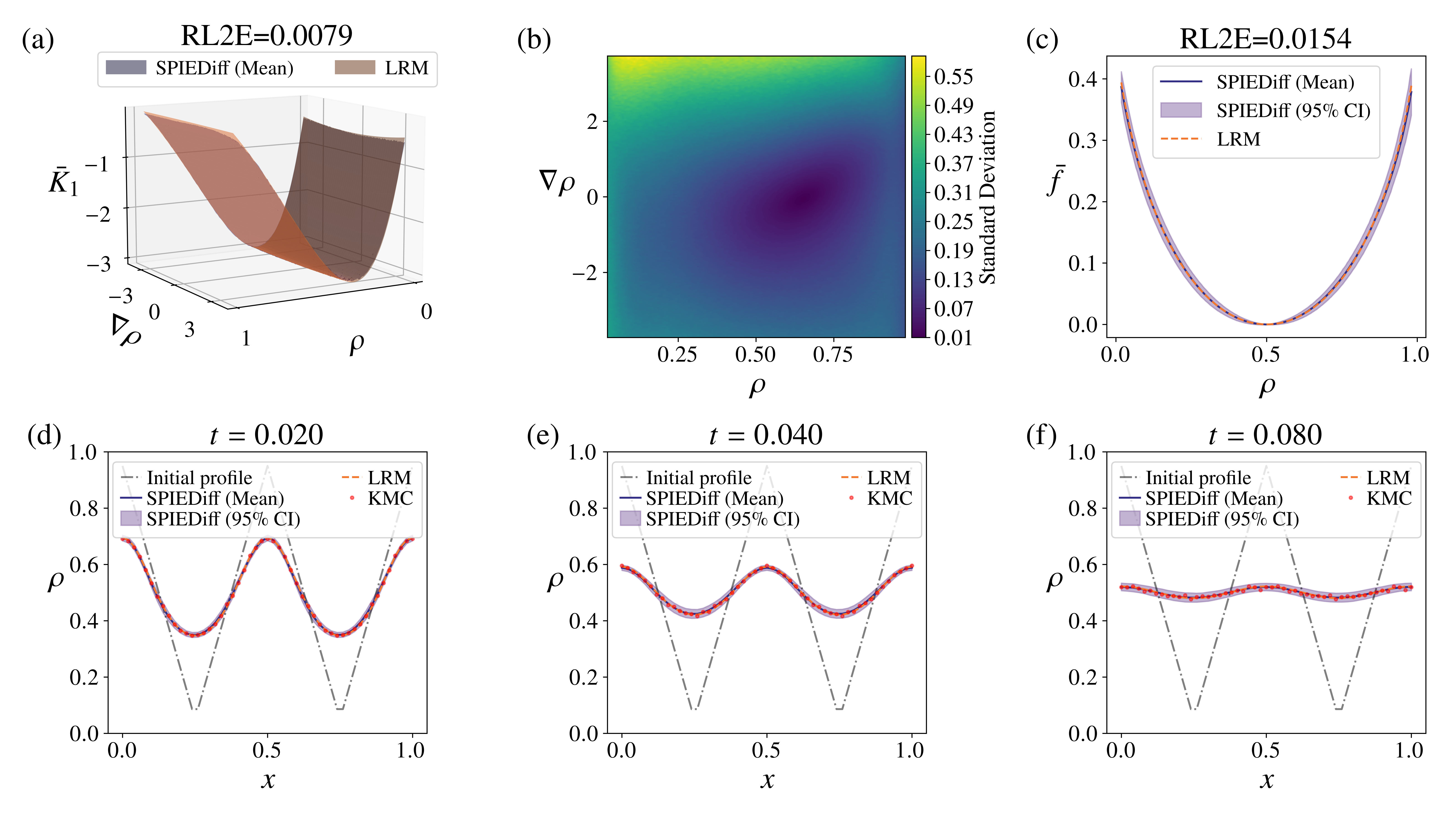}
    \caption{SPIEDiff results compared with the analytical long-range model (LRM) and kinetic Monte Carlo (KMC) simulations for the case of long-range interactions: (a) dissipative operator entry $\bar{K}_1$, (b) its epistemic uncertainty, quantified by SPIEDiff, (c) calibrated free energy density $\bar{f}$, and (d-f) macroscopic evolution snapshots at indicated times (CI denotes confidence interval).}
    \label{fig:LR_R4_all}
\end{figure}

\textbf{Long-range interactions.} Figure \ref{fig:LR_R4_all} presents the SPIEDiff results for the Arrhenius process with long-range interactions, using the full dataset available from Stat-PINNs. 
Mean predictions for the learned $\bar{K}_1$ and $\bar{f}$ closely match the LRM, yielding relative $L^2$ errors (RL2E) of $0.79 \%$ and $1.54 \%$, respectively (Fig.~\ref{fig:LR_R4_all}a, c). Compared to previously reported Stat-PINNs results \cite{huang2025statistical} ($0.59 \%$ for $\bar{K}_1$ and $2.56 \%$ for $\bar{f}$), SPIEDiff exhibits a marginally higher error for $\bar{K}_1$ but achieves notably better accuracy for $\bar{f}$. 
The corresponding estimates of epistemic uncertainty are visualized via a standard deviation map for $\bar{K}_1$ (Fig.~\ref{fig:LR_R4_all}b) and $95\%$ confidence intervals (CI) for $\bar{f}$ (Fig.~\ref{fig:LR_R4_all}c). Furthermore, the macroscopic dynamics predicted by SPIEDiff demonstrate excellent agreement with LRM over time, with the uncertainty bounds consistently encompassing the LRM and KMC data points (Fig.~\ref{fig:LR_R4_all}d-f), indicating reliable predictions.

\textbf{Short-range interactions.} We next evaluate SPIEDiff in the regime of short-range interactions (both strong and weak interactions, using the full training dataset), where the LRM is known to fail.  
Figure \ref{fig:weak_SR_R4_all} presents the results for weak short-range interactions. In this case, while the learned mean free energy density $\bar{f}$ remains visually consistent with the LRM form (Fig.~\ref{fig:weak_SR_R4_all}c), SPIEDiff discovers a distinct kinetic component $\bar{K}_1$ that noticeably deviates from the LRM prediction (Fig.~\ref{fig:weak_SR_R4_all}a). Importantly, the macroscopic dynamics predicted by SPIEDiff align closely with KMC simulations (Fig.~\ref{fig:weak_SR_R4_all}d-f), whereas the LRM predictions exhibit clear discrepancies. SPIEDiff also provides uncertainty quantification shown in Fig.~\ref{fig:weak_SR_R4_all}b for $\bar{K}_1$ and Fig.~\ref{fig:weak_SR_R4_all}c for $\bar{f}$ as well as the macroscopic predictions in Fig.~\ref{fig:weak_SR_R4_all}d-f with the KMC data points appropriately covered. 

Finally, Fig.~\ref{fig:strong_SR_R4_all} demonstrates SPIEDiff's performance in the challenging strong short-range interaction case, where the LRM is known to fail qualitatively. Both the learned $\bar{K}_1$ (Fig.~\ref{fig:strong_SR_R4_all}a) and $\bar{f}$ (Fig.~\ref{fig:strong_SR_R4_all}c) discovered by SPIEDiff differ substantially from the LRM predictions. Especially, SPIEDiff learns a single-well free energy while the LRM erroneously predicts a double-well free energy landscape associated with phase separation. The macroscopic evolution predicted by SPIEDiff is consistent with the KMC simulation results (Fig.~\ref{fig:strong_SR_R4_all}d-f), whereas the LRM predicts incorrect dynamics. Meanwhile, the epistemic uncertainty quantified by SPIEDiff is visualized for $\bar{K}_1$ (Fig.~\ref{fig:strong_SR_R4_all}b), $\bar{f}$ (Fig.~\ref{fig:strong_SR_R4_all}c), and the dynamics (Fig.~\ref{fig:strong_SR_R4_all}d-f) where most of the KMC data points are successfully captured.

\textbf{Low computational cost.} A potential computational bottleneck for SPIEDiff arises from its use of conditional DDPMs as base networks due to the sequential Markovian nature of the DDPM sampler.
This challenge is amplified when thousands of realizations are needed to quantify uncertainty during continuum prediction. Therefore, we replace the original DDPM sampler with the DDIM sampler, introduced in Section \ref{sec:prelim}, to ensure efficient inference and prediction. By employing such a sampler, we drastically reduce the number of required reverse diffusion steps from 50 steps down to just 2 steps in our experiments while observing negligible degradation in prediction accuracy, as listed in Table \ref{tab:ddimSamplerError}. Moreover, SPIEDiff is fully implemented in JAX, which allows rapid execution on hardware accelerators. Table \ref{tab:totalComputCost} provides a comparison of the total computational cost across different models. Since SPIEDiff and Stat-PINNs are trained and used for prediction on a single GPU, the listed time does not consider any parallelization on the hardware. Even accounting for full parallelization, which might reduce the full KMC simulation time to the order of days, the SPIEDiff and Stat-PINNs remain substantially faster for generating macroscopic predictions. We remark that the listed training time for SPIEDiff is the total training time for both the base networks and the epinets. Not only that, the prediction time is the total time for the $2000$ realizations used for the macroscopic prediction, which makes SPIEDiff a computationally efficient framework even compared to Stat-PINNs. 
\begin{table}[htbp]
  \caption{Comparison of total computational costs for SPIEDiff, Stat-PINNs, and full KMC model.}
  \label{tab:totalComputCost}
  \centering
  \begin{tabular}{l c cc cc l}
    \toprule
      & \multicolumn{1}{c}{Train Data} 
      & \multicolumn{2}{c}{SPIEDiff} 
      & \multicolumn{2}{c}{Stat-PINNs} 
      & KMC (full) \\
    \cmidrule(lr){3-4} \cmidrule(lr){5-6}
    \multicolumn{1}{c}{Case} 
      & \multicolumn{1}{c}{Collection} 
      & Train
      & Predict (+ UQ)
      & Train 
      & Predict
      & Runtime \\
    \midrule
    Long-range
      & 55~days
      & 45\,sec
      & 2.5\,min
      & 3.5\,min
      & 1\,min
      & 3500~days \\

    Short-range (weak)
      & 19~days
      & 45\,sec
      & 2.5\,min
      & 3.5\,min
      & 1\,min
      & 875~days \\

    Short-range (strong)
      & 39~days
      & 45\,sec
      & 6.5\,min
      & 3.5\,min
      & 3\,min
      & 625~days \\
    \bottomrule
  \end{tabular}
\end{table}

\begin{figure}[htbp]
    \centering
    \includegraphics[width=1.0\linewidth]{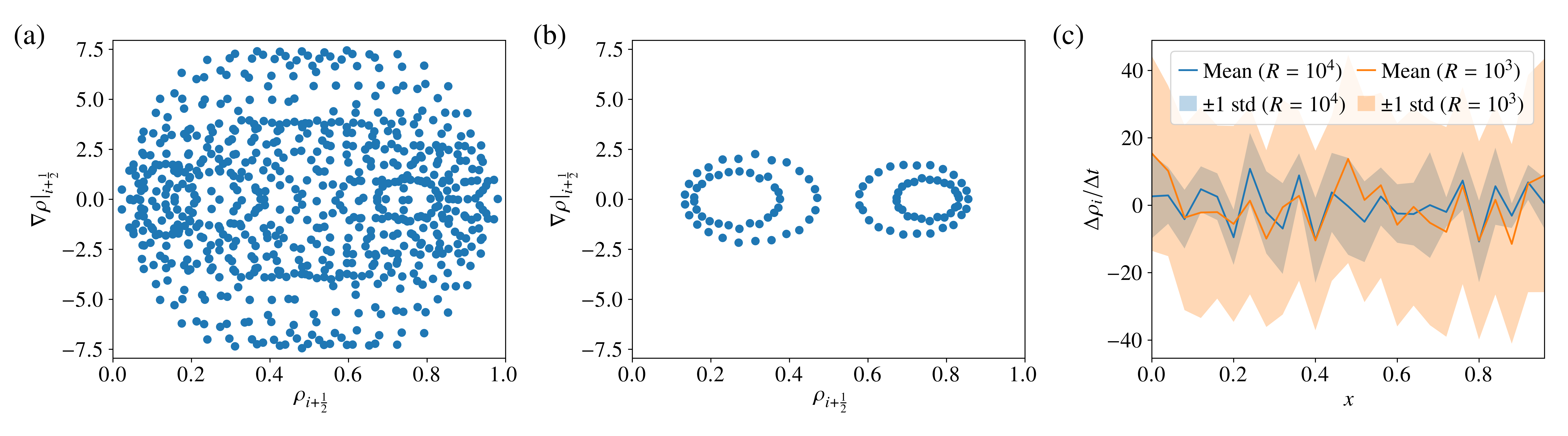}
    \caption{Comparison of the training datasets used for the case of long-range interactions. These are generated from (a) $28$ initial profiles, with $R=10^4$ and (b) the first $4$ initial profiles with $R=10^4$. (c) shows $\Delta \rho_i / \Delta t$ obtained form keeping $28$ initial profiles, but varying $R$ for each profile.}
    \label{fig:datasetCompare}
\end{figure}

\begin{figure}[htbp]
    \centering
    \includegraphics[width=1.0\linewidth]{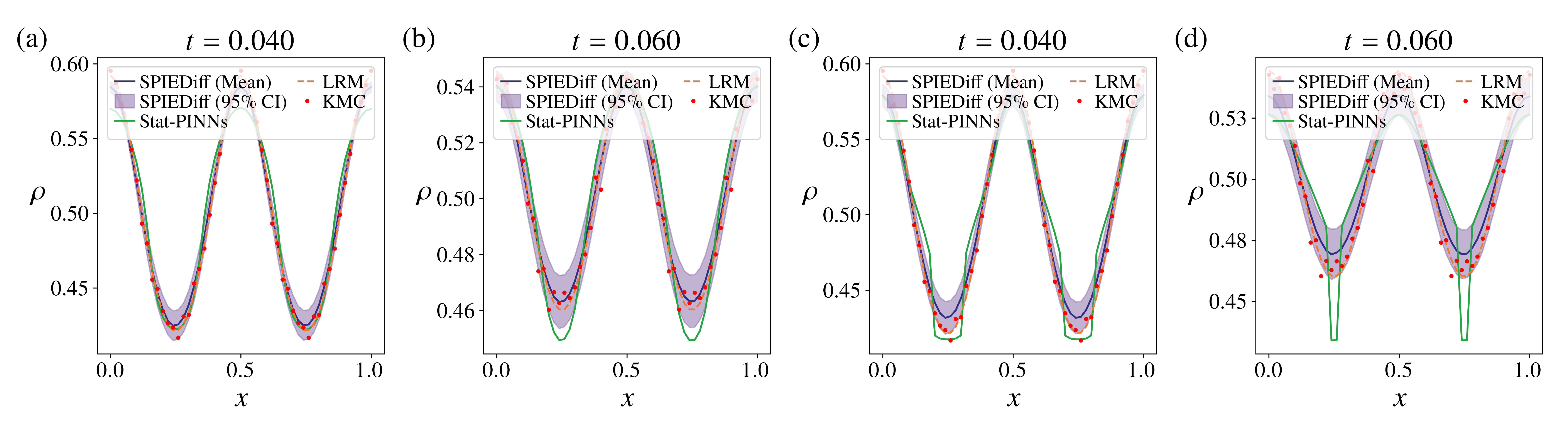}
    \caption{Comparison of SPIEDiff and baseline models (Stat-PINNs, LRM) under challenging data conditions at selected time instances. (a) and (b) show the results for the scarcer data condition at $t = 0.040$ and $t = 0.060$, respectively, while (c) and (d) present the results for the noisier data condition at the same time instances. For SPIEDiff, both mean predictions and $95 \%$ confidence intervals (CI) are shown.}
    \label{fig:scarceNoiseCompare}
\end{figure}

\textbf{Robust to challenging data conditions.} The dominating computational cost for the ML-based frameworks is not due to the model training or prediction, but rather due to the collection of training data via short-time KMC simulations. In all the previous experiments, we used 28 profiles with $R=10^4$ realizations for each profile. An example of the training data space for the case of long-range interactions is shown in Fig.~\ref{fig:datasetCompare}a. Hence, we challenge the SPIEDiff with two difficult data conditions: (i) training with $4$ initial profiles while maintaining $R=10^4$ realizations per profile (scarcer data as shown in Fig.~\ref{fig:datasetCompare}b), and (ii) training with a reduced number of realizations ($R=10^3$) per profile while keeping the original set of initial profiles (noisier data as indicated in Fig.~\ref{fig:datasetCompare}c). For the scarcer data case, the number of training data is reduced from $700$ to $100$, and the data is also inherently noisy. Nonetheless, SPIEDiff shows excellent agreement with the analytical model (LRM) and provides uncertainty bounds that correctly cover most of the KMC data points (Fig.~\ref{fig:scarceNoiseCompare}a,b). In the scenario with noisier data, SPIEDiff provides more accurate macroscopic predictions compared to Stat-PINNs and includes most of the KMC data within the uncertainty bounds. A quantitative comparison is provided in Table ~\ref{tab:compareDataEfficiency}. We note that, although the results presented here already reveal noticeable fluctuations in Stat-PINNs' predictions, Stat-PINNs could completely fail under challenging data conditions without careful fine-tuning. Conversely, SPIEDiff consistently delivers more accurate and stable predictions compared to Stat-PINNs.

\section{Conclusion}
\label{sec:conclusion}
In this work, we introduced SPIEDiff, a framework for learning long-time continuum dynamics and their true underlying thermodynamic structure with quantified epistemic uncertainty from short-time particle simulations. Built upon Stat-PINNs, SPIEDiff addresses the non-uniqueness problem inherent in learning thermodynamic structures from macroscopic data by leveraging fluctuation-dissipation relations from statistical mechanics. This allows for the unique determination of the dissipative operator and free energy functional. SPIEDiff robustly discovers these underlying thermodynamic structures by integrating conditional DDPMs as base networks with epinets for UQ. Our results show that SPIEDiff outperforms Stat-PINNs with enhanced robustness to noisy and scarce data, provides efficient epistemic uncertainty for both learned components and predictions, and can discover the thermodynamics and kinetics for systems with unknown analytical coarse-grained models. 

\textbf{Limitations:} Current limitations include the need for further calibration of the epistemic uncertainty to ensure rigorous statistical coverage and managing sensitivity to hyperparameters and initial profiles used in short-time particle simulations. Future work will also focus on extending SPIEDiff to more complex, higher-dimensional physical systems, including those governed by the full GENERIC structure.

\textbf{Broader impacts:} SPIEDiff contributes to the broader goal of building more robust and trustworthy data-driven models for scientific discovery of thermodynamic models. This work has the potential to accelerate the design and understanding of complex materials, chemical processes, and other multiscale phenomena. 

\begin{ack}
The authors acknowledge support from the US Department of the Army W911NF2310230, and thank Dr. Shenglin Huang for providing the dataset and valuable discussions about the Stat-PINNs framework.
\end{ack}

{\small
\bibliographystyle{unsrtnat}
\bibliography{references}
}


\appendix

\section{Model architecture and training details}
\label{sec:modelArchTrain}
\begin{figure}[htbp]
    \centering
    \includegraphics[width=0.9\linewidth]{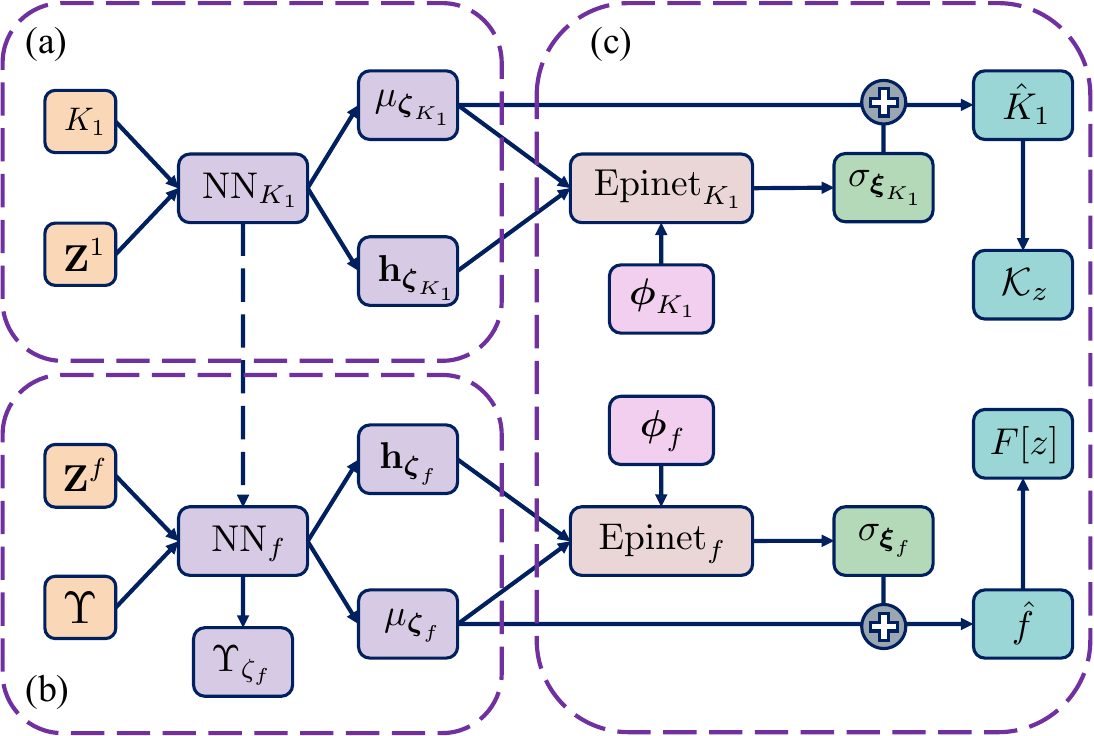}
    \caption{Detailed schematic of the SPIEDiff framework architecture and sequential training procedure. The process involves training base conditional DDPMs ($\mathrm{NN}_{K_1}$ and $\mathrm{NN}_{f}$) followed by corresponding epinets ($\mathrm{Epinet}_{K_1}$ and $\mathrm{Epinet}_{f}$) to learn thermodynamic components with quantified epistemic uncertainty. The training order is: (a)$\rightarrow$(b)$\rightarrow$(c).}
    \label{fig:spiediffDetail}
\end{figure}
\subsection{SPIEDiff model architecture}
\label{sec:modelArch}
The SPIEDiff architecture, depicted schematically in Fig.~\ref{fig:spiediffDetail}, uses a sequential training strategy to learn both the thermodynamic components ($\mathcal{K}_z, F[z]$) of the coarse-grained dynamics and quantify their associated epistemic uncertainty. Initially (Fig.~\ref{fig:spiediffDetail}a, b), two independent base conditional DDPMs, $\mathrm{NN}_{K_1}$ (parameters $\boldsymbol{\zeta}_{K_1}$) and $\mathrm{NN}_f$ (parameters $\boldsymbol{\zeta}_f$), are trained using physics-informed losses (Eqs.~\eqref{eq:lossFuncionK1Base}, \eqref{eq:lossFuncionFEBase}) to learn the off-diagonal term of the discretized dissipative operator ($K_1$) and the free energy density ($f$). The non-positive transformation function to impose the positive semi-definiteness of $\mathcal{K}_z$ is chosen based on previous studies \cite{sivaprasad2021curious, huang2022variational} as $g(K_1)=-K_1-\mathrm{exp}(-5)$ for $K_1 \geq 0$ and $g(K_1)=-\mathrm{exp}(K_1-5)$ when $K_1 < 0$. These trained base networks provide base predictions $\mu_{\boldsymbol{\zeta}_{K_1}}$ and $ \mu_{\boldsymbol{\zeta}_f}$ and features $\mathbf{h}_{\boldsymbol{\zeta}_{K_1}}$ and $\mathbf{h}_{\boldsymbol{\zeta}_f}$. The quantity $\Upsilon$ serves as an auxiliary target variable for the purpose of training the base network $\mathrm{NN}_f$ in the reverse process. In this work, $\Upsilon$ is defined using the macroscopic evolution data, in particular, $\sum_i \langle \gamma_{j}, \gamma_i \rangle \Delta z_i^{(s)} / \Delta t$. The network $\mathrm{NN}_f$ includes a dedicated auxiliary output, $\hat{\Upsilon}$, which is trained via a standard DDPM denoising loss term. We note that the choice of the auxiliary target variable $\Upsilon$ is flexible, and this choice does not significantly affect the performance of SPIEDiff. Subsequently (Fig.~\ref{fig:spiediffDetail}c), with base parameters frozen, the corresponding epinets ($\mathrm{Epinet}_{K_1}$ with parameters $\boldsymbol{\xi}_{K_1}$, $\mathrm{Epinet}_f$ with parameters $\boldsymbol{\xi}_f$) are trained jointly via knowledge distillation losses (Eq.~\eqref{eq:lossFunctionEpinets}). Each epinet utilizes the fixed features ($\mathbf{h}$ via stop-gradient) and a sampled epistemic index ($\boldsymbol{\phi}$) to predict an uncertainty component ($\sigma_{\boldsymbol{\xi}}$). The final predictions are formed by combining the base predictions and the epinet outputs ($\hat{K}_1 = \mu_{\boldsymbol{\zeta}_{K_1}} + \sigma_{\boldsymbol{\xi}_{K_1}}$, $\hat{f} = \mu_{\boldsymbol{\zeta}_f} + \sigma_{\boldsymbol{\xi}_f}$), providing the learned operator $K_z$ and functional $F[z]$ with quantified epistemic uncertainty.

\subsection{Training details}
\label{sec:trainDetails}
Although joint training of the base networks (i.e., for $K_1$ and $f$) is viable, the sequential approach simplifies the fine-tuning of each network component and avoids the potential complexities of adaptive loss weighting schemes. As a result, we adopt the sequential training strategy and set, for simplicity, all loss weights to $1$ within each loss function. Once the base networks are trained, their corresponding epinets are jointly trained. Table \ref{tab:baseHyperparamsMain} and Table \ref{tab:epinetHyperparamsMain} list the hyperparameter settings used in Section \ref{sec:experiments} to train the base networks and epinets of SPIEDiff, respectively. Note that both $\mathrm{Epinet}_{K_1}$ and $\mathrm{Epinet}_{f}$ share the same setting detailed in Table \ref{tab:epinetHyperparamsMain}. For the Stat-PINNs' results of Section \ref{sec:experiments} 
we use the same hyperparameter settings reported in 
\cite{huang2025statistical}. These are summarized in Table \ref{tab:statpinnsHyperparamsMain}. 

\begin{table}[htbp]
  \centering
  \caption{Hyperparameter settings for base networks (used for generating main results).}
  \label{tab:baseHyperparamsMain}
  \begin{tabular}{lcc}
    \toprule
    \textbf{Parameter}           & $\mathrm{NN}_{K_1}$ & $\mathrm{NN}_{f}$ \\
    \midrule
    Training epochs              & 20{,}000            & 20{,}000 \\
    Learning rate                & $10^{-4}$           & $10^{-3}$ \\
    Embedding size               & 2                   & 2 \\
    Time embedding               & Fourier             & Fourier \\
    Number of timesteps          & 50                  & 50 \\
    Beta schedule                & Cosine              & Cosine \\
    Hidden layers                & 3                   & 3 \\
    Nodes per layer              & 50                  & 50 \\
    \bottomrule
  \end{tabular}
\end{table}

\begin{table}[htbp]
  \centering
  \caption{Hyperparameter settings for epinets (used for generating main results).}
  \label{tab:epinetHyperparamsMain}
  \begin{tabular}{lc}
    \toprule
    \textbf{Parameter}           & $\mathrm{Epinet}$ \\
    \midrule
    Training epochs              & 10{,}000 \\
    Learning rate                & $10^{-4}$ \\
    Hidden layers (prior)        & 2 \\
    Nodes per layer (prior)      & 16 \\
    Hidden layers (learnable)    & 2 \\
    Nodes per layer (learnable)  & 16 \\
    Epistemic index size         & 4 \\
    Prior scale                  & 1 \\
    \bottomrule
  \end{tabular}
\end{table}

\begin{table}[htbp]
  \centering
  \caption{Hyperparameter settings for Stat-PINNs (used for generating main results).}
  \label{tab:statpinnsHyperparamsMain}
  \begin{tabular}{lcc}
    \toprule
    \textbf{Parameter}           & $\mathrm{NN}_{K_1}$ & $\mathrm{NN}_{f}$ \\
    \midrule
    Training epochs              & 6{,}000            & 2{,}000 \\
    Learning rate                & $10^{-2}$           & $10^{-2}$ \\
    Hidden layers                & 2                   & 2 \\
    Nodes per layer              & 20                  & 20 \\
    \bottomrule
  \end{tabular}
\end{table}
For the data efficiency experiments discussed in Appendix \ref{sec:dataEfficiencyAnalysis}, both SPIEDiff and Stat-PINNs are individually fine-tuned for each specific data condition to avoid unfair comparisons. In particular, we notice that Stat-PINNs is extremely sensitive to hyperparameter settings when handling these challenging and limited data. In some instances, Stat-PINNs would completely fail in generating continuum predictions due to this instability. None of this instability or failure is observed for SPIEDiff. For comparison purposes, hyperparameters were thus chosen such that such failures would not be observed. The detailed hyperparameter settings for these experiments are provided in Tables \ref{tab:baseHyperparamsScarceNoise}, \ref{tab:epinetHyperparamsScarceNoise}, and \ref{tab:statpinnsHyperparamsScarceNoise}. 

\begin{table}[htbp]
  \centering
  \caption{Hyperparameter settings for base networks under scarcer vs.\ noisier data conditions.}
  \label{tab:baseHyperparamsScarceNoise}
  \begin{tabular}{lcccc}
    \toprule
    \textbf{Parameter}
      & \multicolumn{2}{c}{Scarcer Data}
      & \multicolumn{2}{c}{Noisier Data} \\
    \cmidrule(lr){2-3} \cmidrule(lr){4-5}
      & $\mathrm{NN}_{K_1}$ & $\mathrm{NN}_{f}$
      & $\mathrm{NN}_{K_1}$ & $\mathrm{NN}_{f}$ \\
    \midrule
    Training epochs      & 20{,}000 & 10{,}000 & 20{,}000 & 30{,}000 \\
    Learning rate        & $10^{-4}$ & $10^{-3}$ & $10^{-4}$ & $10^{-4}$ \\
    Embedding size       & 2        & 2        & 2        & 2        \\
    Time embedding       & Fourier  & Fourier  & Fourier  & Fourier  \\
    Number of timesteps  & 50       & 50       & 50       & 50       \\
    Beta schedule        & Cosine   & Cosine   & Cosine   & Cosine   \\
    Hidden layers        & 3        & 3        & 3        & 3        \\
    Nodes per layer      & 50       & 15       & 50       & 50       \\
    \bottomrule
  \end{tabular}
\end{table}

\begin{table}[htbp]
  \centering
  \caption{Hyperparameter settings for epinet under scarcer vs.\ noisier data conditions.}
  \label{tab:epinetHyperparamsScarceNoise}
  \begin{tabular}{lcc}
    \toprule
    \textbf{Parameter} & \multicolumn{2}{c}{Epinet} \\
    & Scarcer Data & Noisier Data \\
    \cmidrule(lr){2-3}
    Training epochs              & 8{,}000  & 8{,}000 \\
    Learning rate                & $10^{-4}$ & $10^{-4}$ \\
    Hidden layers (prior)        & 2        & 2        \\
    Nodes per layer (prior)      & 5        & 16       \\
    Hidden layers (learnable)    & 2        & 2        \\
    Nodes per layer (learnable)  & 10       & 16       \\
    Epistemic index size         & 2        & 4        \\
    Prior scale                  & 1        & 1        \\
    \bottomrule
  \end{tabular}
\end{table}

\begin{table}[htbp]
  \centering
  \caption{Hyperparameter settings for Stat-PINNs under scarcer vs.\ noisier data conditions.}
  \label{tab:statpinnsHyperparamsScarceNoise}
  \begin{tabular}{lcccc}
    \toprule
    \textbf{Parameter}
      & \multicolumn{2}{c}{Scarcer Data}
      & \multicolumn{2}{c}{Noisier Data} \\
    \cmidrule(lr){2-3} \cmidrule(lr){4-5}
      & $\mathrm{NN}_{K_1}$ & $\mathrm{NN}_{f}$
      & $\mathrm{NN}_{K_1}$ & $\mathrm{NN}_{f}$ \\
    \midrule
    Training epochs      & 6{,}000  & 2{,}000  & 6{,}000  & 2{,}500 \\
    Learning rate        & $10^{-2}$ & $10^{-2}$ & $10^{-2}$ & $10^{-2}$ \\
    Hidden layers        & 2        & 2        & 2        & 2        \\
    Nodes per layer      & 20       & 20       & 20       & 20       \\
    \bottomrule
  \end{tabular}
\end{table}
For all the results presented in this work, for both SPIEDiff and Stat-PINNs, we initialize weights following the Xavier uniform distribution \cite{glorot2010understanding}, use Adam \cite{kingma2014adam} as the optimizer, and pick SoftPlus as the activation function. Mini-batching is not used in any of the training.

\section{Inference and continuum predictions}
\label{sec:inferenceContPrediction}
For all the macroscopic results shown in this work, 2000 realizations are generated using the trained epinets, and we compute the mean and standard deviation over the generated realizations. To accelerate inference and continuum prediction, particularly for the efficient generation of multiple samples required for uncertainty quantification, we replace the original DDPM sampler used in SPIEDiff with a DDIM sampler (Section \ref{sec:prelim}). We evaluated the trade-off between sampling speed and accuracy by varying the number of DDIM steps for the long-range interaction case, with results presented in Table \ref{tab:ddimSamplerError}. This analysis confirmed that employing the DDIM sampler with only 2 reverse diffusion steps yields a prediction accuracy comparable to that of the standard, more computationally intensive DDPM sampler (in our case, 50 steps). Therefore, for all experiments reported in this work, the DDIM sampler is configured with 2 timesteps to ensure both rapid prediction and high fidelity.
\begin{table}[htbp]
  \centering
  \caption{Relative $L^2$ error with DDIM sampler compared to DDPM sampler on the long-range interaction case.}
  \label{tab:ddimSamplerError}
  \begin{tabular}{lcc}
    \toprule
    \textbf{Sampler}                & $\bar{K}_1$       & $\bar{f}$ \\
    \midrule
    DDPM sampler ($50$ steps)   & 0.0079            & 0.0154 \\
    DDIM sampler ($10$ steps)   & 0.0079            & 0.0154 \\
    DDIM sampler ($2$ steps)    & 0.0079            & 0.0154 \\
    DDIM sampler ($1$ step)     & 0.0081            & 0.0157 \\
    \bottomrule
  \end{tabular}
\end{table}
When presenting results for the learned free energy, we focus on the case where the free energy density $f$ is assumed to depend only on the local density $\rho$. Consequently, the thermodynamic driving force is $Q=f'(\rho)$. Since the governing dynamics depend only on $Q$ (up to an additive constant absorbed by the operator), the learned free energy density $f(\rho)$ is inherently determined only up to an arbitrary affine function. To remove this ambiguity and allow for consistent comparison between different models or results, we define a calibrated free energy density $\bar{f}(\rho)$:
\begin{equation}
\label{eq:f_calibration}
\bar{f}(\rho) = f(\rho) - f(\rho_{r0}) - f'(\rho_{r1})(\rho - \rho_{r0}) .
\end{equation}
This calibration procedure fixes the integration constants by setting the value of $\bar{f}$ to zero at a reference density $\rho_{r0}$ and its derivative $\bar{f}'$ to zero at a reference density $\rho_{r1}$, i.e., $\bar{f}(\rho_{r0})=0$ and $\bar{f}'(\rho_{r1})=0$. For visualization purposes throughout this work, we consistently select reference points $\rho_{r0} = \rho_{r1} = 0.5$.

Long-time continuum predictions in SPIEDiff are generated by integrating the weak form of the evolution equations using the learned dissipative operator and thermodynamic forces. This system of ODEs for the finite element basis coefficients $\rho_i(t)$ is solved numerically from a given initial condition using a fourth-order Runge-Kutta (RK4) scheme for accuracy and stability with a time step set to $8*10^{-5}$. 

\section{Baseline models}
\label{sec:baselineModels}
\subsection{Long-range analytic model (LRM)}
\label{sec:longRangeModel}
For the Arrhenius-type interacting particle system described in the main text, a macroscopic continuum description exists in the specific case where the interparticle interaction potential $\hat{J}(x)$ is long-range compared to the lattice spacing $\epsilon$. This analytical model, derived in \cite{vlachos2000derivation}, serves as a crucial benchmark for validating SPIEDiff and Stat-PINNs in this work. The macroscopic evolution of the particle density $\rho(x,t)$ is given by the diffusion equation
\begin{equation}
\label{eq:lrmDiffusionEq}
\frac{\partial\rho}{\partial t}=\nabla\cdot\left(m[\rho]\nabla\frac{\delta F[\rho]}{\delta\rho}\right),
\end{equation}
which is a gradient flow of the dimensionless Helmholtz free energy functional $F[\rho]$ given by
\begin{equation}
F[\rho]=-\int \frac{1}{2} \rho(J * \rho)\, \mathrm{d} x+\int\left[\rho \ln \rho+(1-\rho) \ln (1-\rho)\right]\, \mathrm{d} x.
\end{equation}
Here, $J * \rho=\int J(x-\chi) \rho(\chi)\, d \chi$ denotes the convolution between the interaction potential and the density. $m[\rho]$ is the mobility function defined as
\begin{equation}
m[\rho]=D\rho(1-\rho)e^{-J*\rho},
\end{equation}
where $D=d\, e^{-\beta \hat{U}_0}$ is a diffusion coefficient with $d$ being the jump frequency and $\hat{U}_0$ the binding energy. $J(\cdot)=\beta \hat{J}(\cdot)$ is the dimensionless interaction energy. The corresponding thermodynamic driving force $Q=\frac{\delta F[\rho]}{\delta \rho}$, assuming a symmetric interaction potential $J(x)=J(-x)$, is
\begin{equation}
Q=\frac{\delta F}{\delta\rho}=-J*\rho-\ln\left(\frac{1}{\rho}-1\right).
\end{equation}
The dissipative operator consistent with the gradient flow structure $\partial_t \rho=-\mathcal{K}_\rho \frac{\delta F[\rho]}{\delta \rho}$ is then
\begin{equation}
\mathcal{K}_{\rho}=-\nabla\cdot(m[\rho]\nabla) = -\nabla\cdot\left[D\rho(1-\rho)e^{-J*\rho}\nabla\right].
\end{equation}
The spatio-temporal discretization of Eq.~\eqref{eq:lrmDiffusionEq} used to evolve the LRM and the approximations made to evaluate the discretized dissipative operator entries,
follows the methodology detailed in Stat-PINNs \cite{huang2025statistical} (see Appendix G). 

\subsection{Stat-PINNs}
\label{sec:stat-pinns}
As the foundation work of the proposed SPIEDiff, Stat-PINNs here serve as another baseline model. The dissipative operator entry $K_1$ and free energy density $f$ 
are learned through two individual MLPs (parameterized by $\boldsymbol{\theta}_1$ and $\boldsymbol{\theta}_f$, respectively), which are trained sequentially. The loss function for learning the off-diagonal entry $K_1$ is
\begin{equation}
\begin{aligned}
\mathcal{L}_K= & \frac{\lambda_0}{2 N_0} \sum_{s=1}^{N_0}\left\|-K_1\left(\mathbf{Z}_{i^s-1}^{1(s)} ; \boldsymbol{\theta}_1\right)-K_1\left(\mathbf{Z}_{i^s}^{1(s)} ; \boldsymbol{\theta}_1\right)-K_0^{(s)}\right\|^2 \\
& +\frac{\lambda_1}{2 N_1} \sum_{s=1}^{N_1}\left\|K_1\left(\mathbf{Z}_{i^s}^{1(s)} ; \boldsymbol{\theta}_1\right)-K_1^{(s)}\right\|^2,
\end{aligned}
\end{equation}
where $\lambda_0$ and $\lambda_1$ are the loss weights determined by the neural kernel tangent method \cite{wang2022and}. And the loss function for training $f$ is 
\begin{equation}
\mathcal{L}_f=\frac{1}{2 N_f} \sum_{s=1}^{N_f} \frac{\left\|\sum_i\left\langle\gamma_j, \gamma_i\right\rangle \Delta z_i^{(s)} / \Delta t+\sum_i\left\langle\gamma_j, \mathcal{K}_{z^{(s)}} \gamma_i\right\rangle Q\left(\mathbf{Z}_i^{f(s)} ; \boldsymbol{\theta}_f\right)\right\|^2}{\left(\sigma_{E q, j^s}^{(s)}\right)^2}.
\end{equation}
Here, $\left(\sigma_{Eq, j^s}^{(s)}\right)^2$ represents the stochastic noise variance at point $x_j$ for sample $s$. This term, inspired by \cite{dietrich2023learning}, is used to appropriately weight the PDE residuals according to their expected noise levels and is indispensable to Stat-PINNs (we remark that this is not the case for SPIEDiff, where such noise variance is not used). However, since such a term is defined as $\sigma_{E q, j}^2=2 \epsilon\left\langle\gamma_j, \mathcal{K}_z \gamma_j\right\rangle /(R \Delta t)$, if the operator is not well-learned, the error could compromise the subsequent learning of the free energy density. This sensitivity and risk of error propagation are exacerbated under challenging training conditions, such as when data gets scarcer or noisier.

\section{Short-range interactions results}
\label{sec:shortRangeResults}
\begin{figure}[htbp]
    \centering
    \includegraphics[width=1.0\linewidth]{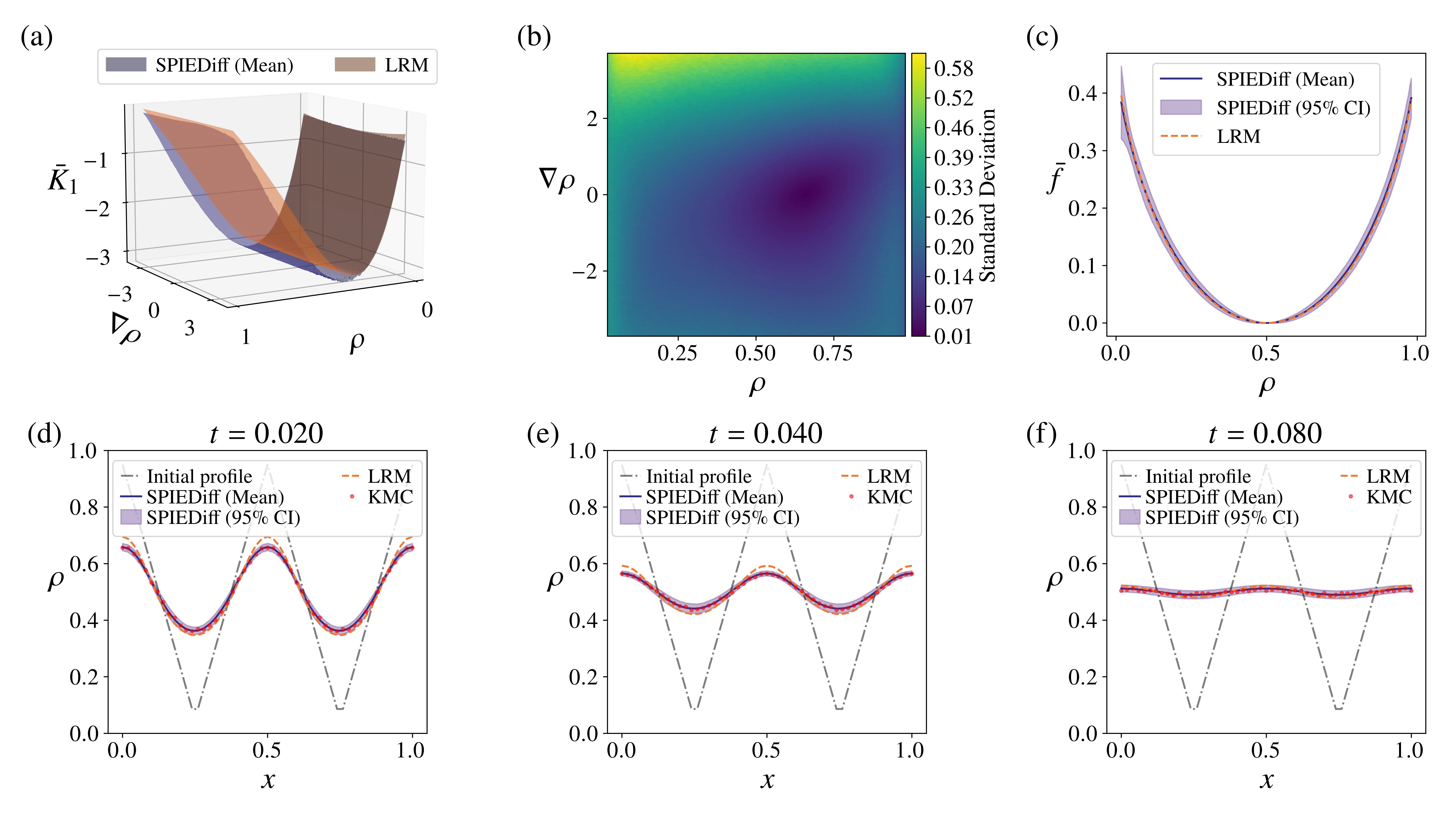}
    \caption{SPIEDiff results compared with the long-range model (LRM) and kinetic Monte Carlo (KMC) simulations of an Arrhenius particle process with weak short-range interactions: (a) dissipative operator entry $\bar{K}_1$, (b) its epistemic uncertainty, quantified by SPIEDiff, (c) the calibrated free energy density $\bar{f}$, and (d-f) macroscopic evolution snapshots at indicated times. $95\%$ confidence interval ($\text{CI}$) represents the quantified epistemic uncertainty by SPIEDiff.}
    \label{fig:weak_SR_R4_all}
\end{figure}

\begin{figure}[htbp]
    \centering
    \includegraphics[width=1.0\linewidth]{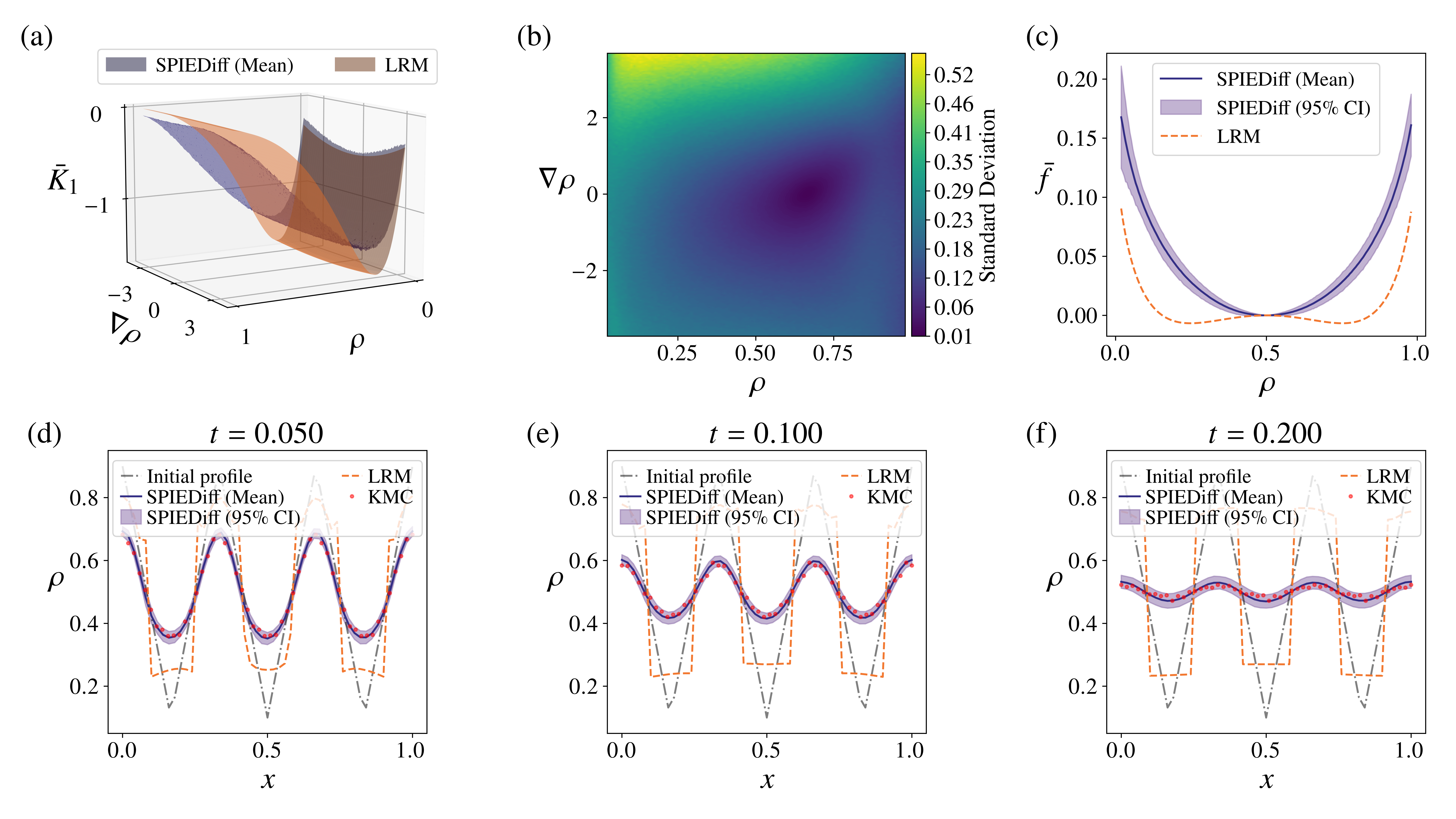}
    \caption{SPIEDiff results compared with the long-range model (LRM) and kinetic Monte Carlo (KMC) simulations of an Arrhenius particle process with strong short-range interactions: (a) dissipative operator entry $\bar{K}_1$, (b) its epistemic uncertainty, quantified by SPIEDiff for $\bar{K}_1$, (c) the calibrated free energy density $\bar{f}$, and (d-f) macroscopic evolution snapshots at indicated times. $95\% $ confidence interval ($\text{CI}$) represents the quantified epistemic uncertainty by SPIEDiff.}
    \label{fig:strong_SR_R4_all}
\end{figure}

In this section, we present the results for the Arrhenius particle process in the regime of short-range interactions, for both weak (see Fig.~\ref{fig:weak_SR_R4_all}) and strong interactions (see Fig.~\ref{fig:strong_SR_R4_all}). As discussed in Section \ref{sec:experiments} and shown in the figures, the LRM is not applicable in these cases. To the best of the authors' knowledge, no macroscopic analytical model is known for these cases. The results of SPIEDiff are thus compared to the computationally intensive full KMC simulations. SPIEDiff not only closely matches the KMC results but also produces uncertainty bounds that cover most of the KMC data points.

\section{Data generation and preparation}
\label{sec:dataGenPrepare}
The training data for SPIEDiff, as well as the particle data used for validating the long-time continuum predictions, are generated from simulations of an Arrhenius-type interacting particle process. These are KMC simulations performed using the Bortz–Kalos–Lebowitz (BKL) algorithm \cite{bortz1975new}. For the specific examples presented (long-range, weak short-range, and strong short-range interactions), we simulate a particle system on a one-dimensional lattice containing 2000 bins, with its state represented macroscopically using 25 finite element shape functions. Data generation commences from 28 distinct initial cosine density profiles listed in Table \ref{tab:initProfiles}. To extract the data needed for learning the dissipative operator, fluctuation measurements are obtained from $R=10^4$ KMC realizations per initial profile, analyzed over $N_h=10$ short consecutive time intervals of length $h$. For learning the free energy density, the average macroscopic evolution is computed over a time interval $\Delta t$ (where $\Delta t \gg h$), also averaged across $R=10^4$ realizations per initial profile. The complete settings of the KMC simulations (including the values for $h$, $\Delta t$, and the equilibration time $t_{eq}$), and the specifics of the Arrhenius interaction potentials for each case study are identical to those used in the original Stat-PINNs paper. For these exhaustive details, interested readers are referred to Appendix H of the paper by Huang et al. \cite{huang2025statistical}. The KMC datasets used for Stat-PINNs are publicly available at \url{https://github.com/celiareina/Stat-PINNs}. These same datasets are used in this work.

\begin{table}[htbp]
  \caption{Parameters for the 28 initial KMC simulation profiles used to generate the training dataset, defined by $\rho(x)=\rho_{\mathrm{ave}}-A\cos(4 \pi fx)$.}
  \label{tab:initProfiles}
  \centering
  \begin{tabular}{@{} c c c c @{}} 
    \toprule
    Label & $f$ 
          & $A$ 
          & $\rho_{\mathrm{ave}}$ \\
    \midrule
    1–7   
      & 1 
      & $0.05 + 0.03\,i,\;i=1,\dots,7$ 
      & $0.5 - (-1)^i \times 0.6\,(0.5 - A),\;i=1,\dots,7$ \\

    8–14  
      & 1 
      & $0.31$ 
      & $0.272 + 0.057\,i,\;i=1,\dots,7$ \\

    15–21 
      & 2 
      & $0.04 + 0.03\,i,\;i=1,\dots,7$ 
      & $0.5 + (-1)^i \times 0.84\,(0.5 - A),\;i=1,\dots,7$ \\

    21–28 
      & 2 
      & $0.29$ 
      & $0.272 + 0.057\,i,\;i=1,\dots,7$ \\
    \bottomrule
  \end{tabular}
\end{table}

To specifically evaluate the enhanced robustness of SPIEDiff, we derived two challenging datasets from the original long-range interaction dataset (Fig.~\ref{fig:datasetCompare}a). The first, termed the `scarcer dataset', was created by selecting data generated from only the first 4 (out of the 28) initial cosine profiles, significantly reducing the diversity and size of the training samples (compare Fig.~\ref{fig:datasetCompare}a and \ref{fig:datasetCompare}b). The second, the `noisier dataset', was constructed by reducing the number of KMC realizations used for averaging from $R=10^4$ down to $R=10^3$ for each of the 28 initial profiles before performing coarse-graining. This reduction substantially increases the noise in the derived quantities, such as the macroscopic rate $\Delta \rho_i / \Delta t$ used for learning the free energy density $f$, as depicted in Fig.~\ref{fig:datasetCompare}c.

\section{Data efficiency analysis}
\label{sec:dataEfficiencyAnalysis}
In this section, we present additional results for the two challenging data conditions (scarcer and noisier) that have been discussed in Section \ref{sec:experiments}. The quantitative comparison is shown in Table \ref{tab:compareDataEfficiency} where values represent the relative $L^2$ error for the two learned components: the off-diagonal kinetic function ($\bar{K}_1$) and the calibrated free energy density $\bar{f}$, evaluated against the analytical LRM results. To evaluate the accuracy of the macroscopic dynamics, we use the relative $L^2$ error computed over space at each prediction timestep and report the maximum value.
\begin{table}[htbp]
  \centering
  \caption{Comparison of SPIEDiff and Stat-PINNs' performance under varying data conditions.}
  \label{tab:compareDataEfficiency}
  \begin{tabular}{l l c c}
    \toprule
    Case                                          & Quantity         & SPIEDiff        & Stat‐PINNs         \\
    \midrule
    \shortstack[l]{Scarcer Data\\(First 4 profiles)}   
                                                  & $\bar{K}_1$ RL2E     & 0.0063            & \textbf{0.0048}  \\
                                                  & $\bar{f}$ RL2E       & \textbf{0.0540}   & 0.2181           \\
                                                  & Dynamics Max.~RL2E        & \textbf{0.0120}   & 0.0764           \\
    \midrule
    \shortstack[l]{Noisier Data\\($R=10^3$)}
                                                  & $\bar{K}_1$ RL2E     & 0.0089            & \textbf{0.0070}  \\
                                                  & $\bar{f}$ RL2E      & \textbf{0.0456}   & 0.0866           \\
                                                  & Dynamics Max.~RL2E        & \textbf{0.0173}   & 0.0350           \\
    \bottomrule
  \end{tabular}
\end{table}

\begin{figure}[htbp]
    \centering
    \includegraphics[width=1.0\linewidth]{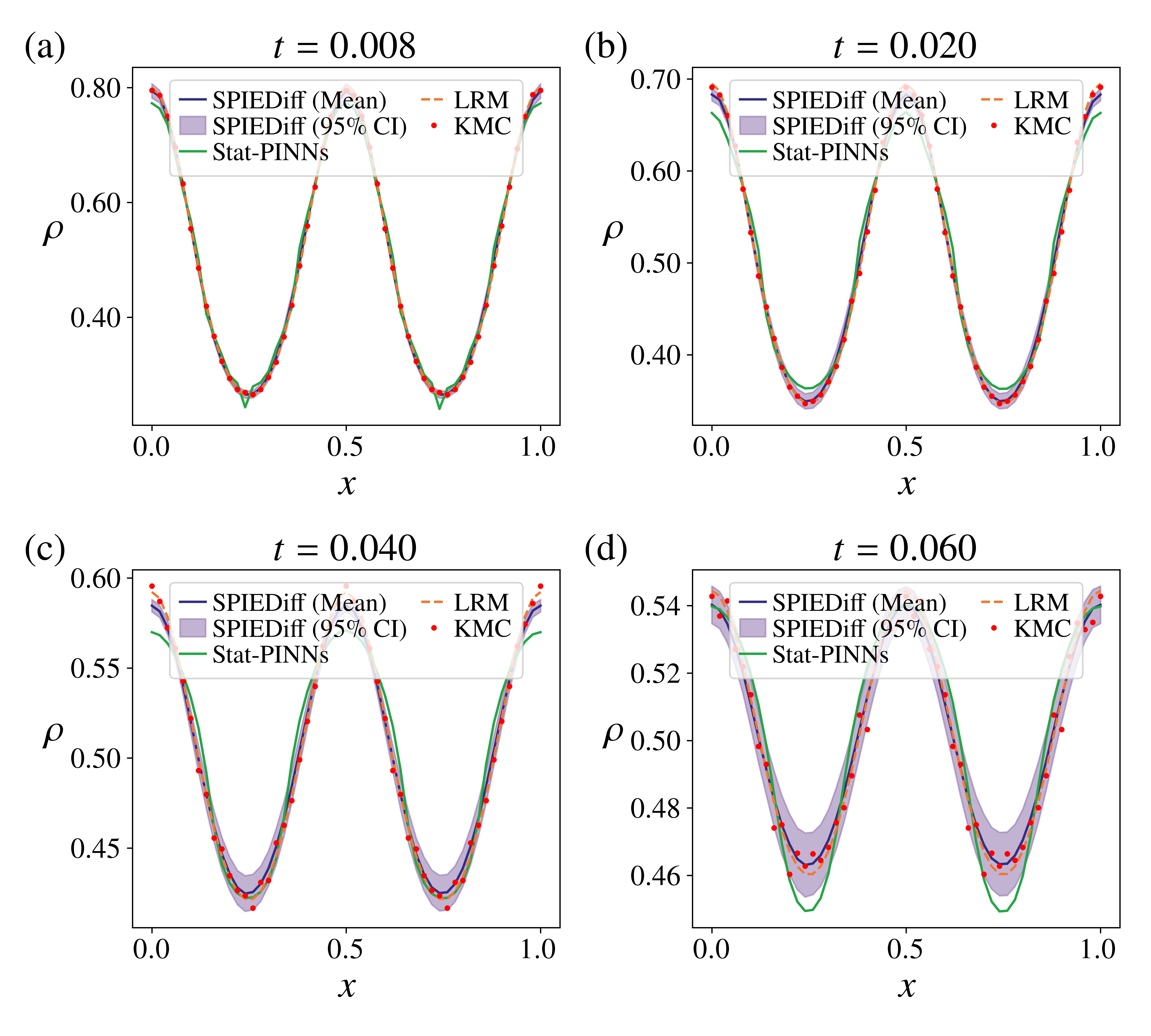}
    \caption{Evaluation of SPIEDiff's predictive accuracy and uncertainty quantification compared to baseline models (LRM and Stat-PINNs) under a scarcer data scenario, validated against KMC simulations. The figure shows snapshots of the macroscopic density $\rho(x,t)$ at four time points: (a) $t=0.008$, (b) $t=0.020$, (c) $t=0.040$, and (d) $t=0.060$. For SPIEDiff, both mean predictions and $95\%$ confidence intervals (CI) are shown.}
    \label{fig:LR_R4_4IP_all}
\end{figure}

\begin{figure}[htbp]
    \centering
    \includegraphics[width=1.0\linewidth]{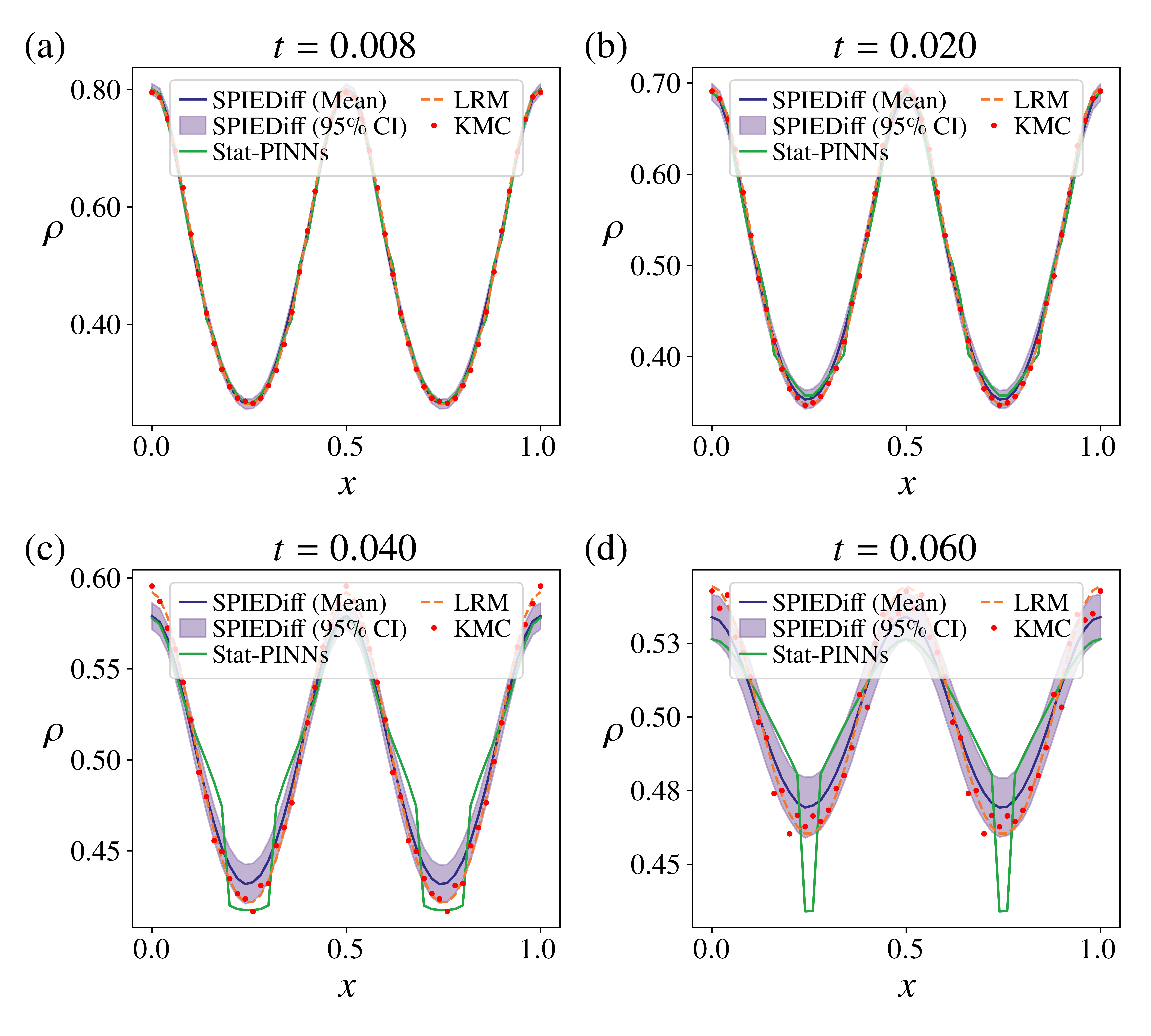}
    \caption{Evaluation of SPIEDiff's predictive accuracy and uncertainty quantification compared to baseline models (LRM and Stat-PINNs) under a noisier data scenario, validated against KMC simulations. The figure shows snapshots of the macroscopic density $\rho(x,t)$ at four time points: (a) $t=0.008$, (b) $t=0.020$, (c) $t=0.040$, and (d) $t=0.060$. For SPIEDiff, both mean predictions and $95\%$ confidence intervals (CI) are shown.}
    \label{fig:LR_R3_all}
\end{figure}

In Figs.~\ref{fig:LR_R4_4IP_all} and \ref{fig:LR_R3_all}, we show a more detailed temporal evolution of the density profiles given by SPIEDiff compared to the baseline models (LRM and Stat-PINNs) under the challenging data conditions (a reduced version was shown in Fig.~\ref{fig:scarceNoiseCompare}). As discussed in Section \ref{sec:experiments}, Stat-PINNs can exhibit fluctuations or even fully fail for these reduced data conditions. SPIEDiff, however, provides reasonably good predictions and remains robust overall.


\section{Hardware and implementation}
\label{sec:hardwareImplement}
KMC simulations of the Arrhenius process are used for two purposes: (i) to generate short-time trajectory data for training the machine learning frameworks, and (ii) to produce long-time evolution data for validating the discovered continuum equations. These particle simulations were implemented in C++ and run on Intel(R) Xeon(R) E5-2683 v4 CPUs @ 2.10 GHz. SPIEDiff and Stat-PINNs are implemented in Python using JAX/Flax. Other standard libraries such as Numpy, Scipy, and Matplotlib are used for data processing and visualizations. The training, inference, and prediction of both ML models are on a single NVIDIA RTX A6000 GPU.

\end{document}